# Formal Concept Analysis with Three Types of Negation

Zhenghua Pan

School of Mathematics and Data Science, Jiangnan University, Wuxi 214122, China

panzh@jiangnan.edu.cn

**Abstract:** Classic Formal Concept Analysis (FCA) primarily focuses on the positive relationships between objects and attributes and does not have mechanisms for handling negation. To overcome this limitation, we introduce three types of negation concepts (contradictory negation, opposite negation, intermediary negation) into FCA. Based on the set SCOI and logic LCOI+PLCOI with these three types negation, we define formal context, Galois connection operators, formal concept and concept lattice with three types of negation, this leads to the proposal of a "FCACOI: Formal Concept Analysis with contradictory negation, opposite negation and intermediary negation".

For the reasoning in FCACOI, this paper focuses on attribute implication reasoning. Based on the logic LCOI+PLCOI and its semantics, we introduce the notion of "ICOI-entailment" as "semantic implication" for attribute implication reasoning in FCACOI. Through ICOI-entailment, a connection is established between attribute implication reasoning in FCACOI and inference in the logic LCOI+PLCOI, it indicate that formally proven inference rules (theorems) in LCOI+PLCOI are valid in the attribute implication reasoning of FCACOI, LCOI+PLCOI provides a logical foundation for attribute implication reasoning in FCACOI. To illustrate the capability of attribute implication reasoning in FCACOI, we discuss its application in a concrete example. Moreover, we explore attribute reduction of the formal context in FCACOI, propose two research frameworks for attribute reduction from different perspectives, and compare their characteristics.

We believe that Formal Concept Analysis FCACOI with three types of negation is not only an extension of classical FCA that remedies its shortcomings in handling negation relations, but also a necessary response to the demands of complex negation semantics. Based on richer logic and semantics, FCACOI elevates FCA from a theory that "describes affirmations" to one that can "describe affirmations and its contradiction (either this or that), opposition (extreme negation) and intermediary (transitional states between oppositions)".

This work was supported by the National Natural Science Foundation of China [60575038, 60973156, 61375004]; the State Key Laboratory for Novel Software Technology, Nanjing University, P.R. China [KFKT2020B01]

## 1. Introduction

Negation is a complex natural language phenomenon and also an important basic concept in knowledge [1]. Broadly speaking, negation connects an expression E with another expression whose meaning is, in some sense, opposed to that of E [2]. Therefore, the key challenge in understanding negation is to identify the meaning that is in some way opposed to E—this is a semantically complex and highly vague task [3]. Research on handling negation has not yet solved computational processing issues due to its complexity, the various linguistic forms it can take, and the different ways it affects words within its scope. This has made it a research hotspot across different knowledge domains [4].

Formal Concept Analysis (FCA), proposed by the German mathematician Wille, is a widely used method for data analysis and knowledge processing. FCA aims to systematically identify and characterize the conceptual structures within data through formal contexts, formal concepts, and concept lattices, thereby revealing the inherent hierarchical relationships and attribute implications within the data[5,6]. In extended studies of FCA, the combination with different theories is often employed to address the limitations of classical binary logic when dealing with uncertainty, incomplete information, and dynamic data[7,13]. Examples include "Fuzzy Formal Concept Analysis"[14]; "Three-way Formal Concept Analysis"[15]; and "Probabilistic Formal Concept Analysis" [16], among others. FCA and its extensions have been widely applied in many fields such as structured knowledge mining, knowledge representation and reasoning, as well as machine learning and artificial intelligence [7, 13].

However, classical FCA mainly focuses on the positive relationships between objects and attributes. Neither its original definition nor its standard form involves negation; it contains only affirmative information and no negative information. In other words, FCA does not have mechanisms to handle negation and related relationships.

Regarding negation in knowledge and information, we have proposed three distinct types of negation at the conceptual level in references [8] and [9]: contradictory negation, opposite negation and intermediary negation, and constructed the set SCOI and the logic LCOI+PLCOI that incorporate these three types of negation. To enable FCA to handle different types of negation information, in this paper, we introduce the concepts of contradictory negation, opposite negation and intermediary negation into FCA. Based on the set SCOI and the logic LCOI+PLCOI, we define formal contexts, Galois connection operators, formal concepts and concept lattices with three types of negation, and propose a formal concept analysis with contradictory negation, opposite negation and intermediary negation, abbreviated as FCACOI.

For the reasoning in formal concept analysis FCACOI, this paper focuses on attribute implication reasoning. Based on the semantics of the logic LCOI+PLCOI, we introduce the notion of "ICOI-entailment" as the "semantic implication" for attribute implication reasoning in FCACOI. Through ICOI-entailment, a connection is established between attribute implication reasoning in FCACOI and inference in the logic LCOI+PLCOI. This demonstrates that formally proven inference rules in LCOI+PLCOI are valid in the attribute implication reasoning of FCACOI, and that LCOI+PLCOI provides a logical foundation for attribute implication reasoning in FCACOI. To illustrate the capability of attribute implication reasoning in FCACOI, we discuss its application in a concrete example. Moreover, we explore attribute reduction of the formal context in FCACOI, propose two research frameworks for attribute reduction from different perspectives, and compare their characteristics.

This paper introduces the concepts of contradictory negation, contrary negation and intermediary negation into FCA. Based on the set SCOI and the logic LCOI+PLCOI, it studies and proposes a formal concept analysis FCACOI with three types of negation, aiming to enable FCA to handle different kinds of negation information. To the best of our knowledge, this work is unprecedented. The main contributions of this paper are as follows:

1. Under the framework of the set SCOI and the logic LCOI+PLCOI, formal contexts, formal concepts, and

This work was supported by the National Natural Science Foundation of China [60575038, 60973156, 61375004]; the State Key Laboratory for Novel Software Technology, Nanjing University, P.R. China [KFKT2020B01] concept lattices with contradictory negation, contrary negation, and intermediary negation are proposed.

2. By introducing the definition of the notion "ICOI-entailment", the connection between attribute implication reasoning in FCACOI and formal inference in the logic LCOI+PLCOI is established. The formal inference relations (theorems) proven in LCOI+PLCOI are valid in the attribute implication reasoning of FCACOI. The attribute implication reasoning of FCACOI is logically self-consistent.

3. By applying the attribute implication reasoning of FCACOI in a concrete example, it is demonstrated that FCACOI exhibits stronger reasoning capabilities than conventional FCA.

4. For attribute reduction of the formal context FCCOI of in FCACOI, two attribute reduction research frameworks based on different perspectives are proposed and their characteristics are compared.

This paper is organized as follows: Section 2 briefly introduces the three types of negation and their characteristics. Section 3 outlines the definitions of the set SCOI and logic LCOI+PLCOI with three kinds of negation. We propose a continuous-value semantics for logic LCOI+PLCOI with a truth value domain of [0, 1] and discuss the meta-logical properties of LCOI+PLCOI. In Section 4, formal contexts, Galois connection operators, formal concepts, concept lattices, and formal concept analysis FCACOI with three types of negation are introduced. Section 5 discusses the relationship between attribute implication reasoning in FCACOI and the logic LCOI+PLCOI, as well as the applications of attribute implication reasoning in FCACOI. Section 6 addresses attribute reduction of the formal context FCCOI as well as its research approaches and methods. Section 7 summarizes the main conclusions of this paper and future work.

## 2. Three types of negation in concepts and their characteristics

From the standpoint of artificial intelligence, a fundamental issue in knowledge representation and reasoning is the study and handling of concepts [10]. The logical approach to determining a concept involves identifying its intension and extension, thereby enabling the categorization of the concept [11]. In references [8] and [9], we distinguish between "clear concept" and "fuzzy concept" at the conceptual level, fully understanding the "contradictions" and "oppositions" within the concepts, thereby proposing that there are three different types of negation in the concepts: contradictory negation, opposite negation, and intermediary negation. In this section, we briefly outline these three different types of negation and their characteristics.

The following three different forms of negation exist in both clear and fuzzy concepts:

(1) *Contradictory Negation*. For a species concept under a genus concept, another species concept that has a contradictory relationship with it constitutes a form of negation. We refer to this type of negation as "contradictory negation". In this form of negation, the intensions (connotations) of the two species concepts mutually negate each other, the extensions (denotations) are mutually exclusive (either one or the other), and the sum of the extensions equals the extension of the genus concept. For example, for the species concept 'positive integer' under the genus concept "integer", another species concept 'non-positive integer' is its contradictory negation. For the two species concepts 'daytime' and 'non-daytime' under the genus concept "one day", the latter is the contradictory negation of the former. From this, it can be known that the negation in classical logic is precisely this kind of negation.

(2) *Opposite Negation*. For a species concept under a genus concept, another species concept that has an oppositional relationship with it constitutes another form of negation. We refer to this type of negation as "opposite negation". In this form of negation, the intensions of the two species concepts mutually negate each other and exhibit the greatest difference in intension, but their extensions are not mutually exclusive (not either-or), and the sum of their extensions is less than the extension of the genus concept. For example, for the species concept 'positive integer' under the genus concept "integer", another species concept 'negative integer' is its opposite negation. For the two species concepts 'daytime' and 'night' under the genus concept "one day", the latter is the opposite negation of the former.

(3) *Intermediary Negation*. The intermediary concept between opposite concepts constitutes a (weak) form of negation of the opposite concepts. We refer to this type of negation as "intermediary negation". In this form of negation, the opposite concepts transition through the intermediary concept and the sum of their extensions equals the extension of the genus concept. For example, under the genus concept "integer", between the two opposing species concepts 'positive integer' and 'negative integer', the concept "zero" is their intermediary negation. Under the genus concept "one day", between the two opposing species concepts 'daytime' and 'night', the concepts "dusk" and "dawn" are their intermediary negation.

From the meanings of the three types of negations mentioned above, the contradictory negation is the traditional negation, the opposite negation can be referred to as strong negation, and intermediary negation as weak negation.

We must point out that in reality there are cases where "contradiction" and "opposition" appear identical. Such situations should be understood as contradiction rather than opposition. For example, under the genus concept of 'real numbers', the species concepts 'rational numbers' and 'irrational numbers' (i.e., non-rational numbers) are both contradictory and oppositional. However, since there is no "intermediary" between rational and irrational numbers, they are not oppositional concepts.

To fully understand the meaning of the above three kinds of negation, we further discuss their characteristics in terms of both the intension of the concepts as well as their extensional relations.

(1) Contradictory Negation in Clear Concepts (CNC)

Characteristics of CNC: Extensions are clear, either this or that, and the sum of extensions is equal to the extension of the genus concept.

For example, the positive integer and non-positive integer under the genus concept of "integer" are clear concepts, while the non-positive integer is the contradictory negation of positive integer. The diagram illustrating the extensional relationship between them is shown below (Figure 1).

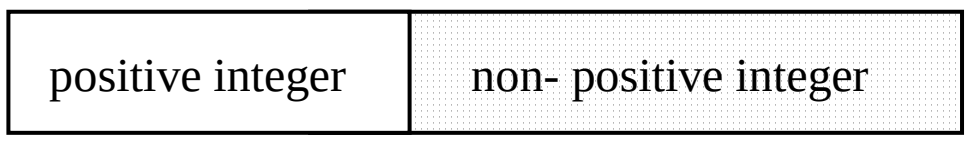

**Fig. 1** The extensional relationship between positive integer and non-positive integer

(2) Opposite Negation in Clear Concepts (ONC)

Characteristics of ONC: Extensions are clear, not "either this or that", and the sum of the extensions is less than the extension of the genus concept.

For example, the positive integer and negative integer under the genus concept of "integer" are clear concepts, while the negative integer is the opposite negation of positive integer. The diagram illustrating the extensional relationship between them is shown below (Figure 2).

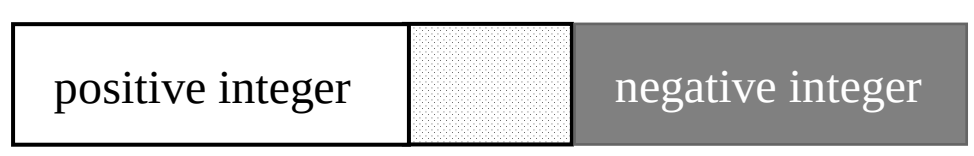

**Fig. 2** The extensional relationship between positive integer and negative integer

(3) Intermediary Negation in Clear Concepts (INC)

Characteristics of INC: Extensions are clear, opposing sides transition to each other through 'intermediaries', and the sum of the extensions equals the extension of the genus concept.

For example, zero, positive integer and negative integer are clear concepts under the genus concept of "integer". Zero serves as an 'intermediary' between positive integer and negative integer, and it is the intermediary negation of both positive integer and negative integer. The diagram illustrating the extensional relationship between them is shown below (Figure 3).

positive integer | zero | negative integer

**Fig. 3** The extensional relationship between positive integers, negative integers, and zero.

(4) Contradictory Negation in Fuzzy Concepts (CNF)

Characteristics of CNF: Extensions are not clear, either this or that, and the sum of extensions is equal to the extension of the genus concept.

For example, the daytime and non-daytime under the genus concept of "day" are fuzzy concepts, while the non-daytime is the contradictory negation of daytime. The diagram illustrating the extensional relationship between them is shown below (Figure 4).

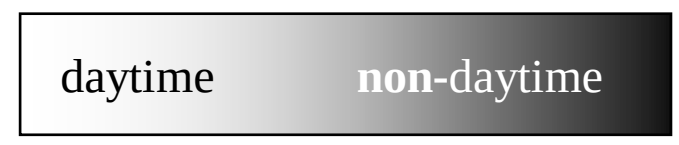

**Fig. 4** The extensional relationship between daytime and non-daytime

(5) Opposite Negation in Fuzzy Concepts (ONF)

Characteristics of ONF: Extensions are not clear, not "either this or that", and the sum of the extensions is less than the extension of the genus concept.

For example, the daytime and night under the genus concept of "day" are fuzzy concepts, while the night is the opposite negation of daytime. The diagram illustrating the extensional relationship between them is shown below (Figure 5).

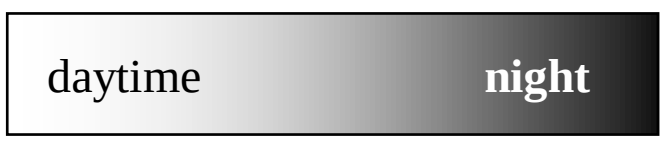

**Fig. 5** The extensional relationship between daytime and night

(6) Intermediary Negation in Fuzzy Concepts (INF)

Characteristics of IFC: Extensions are not clear, opposing sides transition to each other through 'intermediaries', and the sum of the extensions equals the extension of the genus concept.

For example, dusk, daytime and night are fuzzy concepts under the genus concept of "day". Dusk serves as an 'intermediary' between daytime and night, and it is the intermediary negation of both daytime and night. The diagram illustrating the extensional relationship between them is shown below (Figure 6).

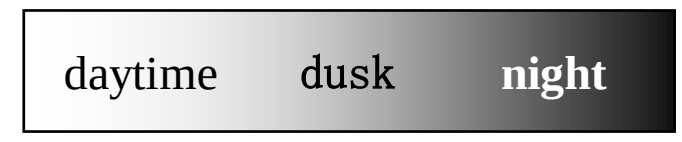

**Fig. 6.** The extensional relationship between daytime, night, and dusk

It is evident that, since concepts are the most fundamental components of knowledge, these three different negations also exist within knowledge itself.

## 3. Set and logic with three kinds of negation

For the three kinds of negation present in the aforementioned concepts, in order to establish a mathematical foundation that can fully reflect them along with their properties, relationships, and laws, we proposed a set SCOI and logic LCOI+PLCOI with contradictory negation, opposite negation and intermediary negation that takes clear and fuzzy entities as the research objects [8, 9].

In this section, we introduce the basic concepts and main conclusions of SCOI and LCOI+PLCOI, and propose a continuous-valued semantics for LCOI+PLCOI with a truth value range of [0, 1]. Under this semantics, the soundness theorem for LCOI+PLCOI is proved.

## 3.1 Set with contradictory negation, opposite negation and intermediary negation

We use symbols ¬, ╕, and ~ to represent 'contradictory negation', 'opposite negation', and 'intermediary negation', respectively.

**Definition 1**. Let $U$ be universe of discourse, $\lambda\in(0, 1)$. Mapping $f: U \to [0, 1]$ confirms a set $A$ on $U$, call $f$ the membership function of $A$, and $f(x)$ the membership degree of $x$ to $A$ (denoted as $A(x)$).

(1) If $A$ is a fuzzy set, then

(*i*). Mapping $f^{╕}: \{A(x) \mid x\in U\} \to [0, 1]$ confirms a fuzzy set $A^{╕}$ on $U$, $A^{╕}(x) = f^{╕}(A(x)) = 1- A(x)$. Call $A^{╕}$ the opposite negation set of $A$.

(*ii*). Mapping $f^{\sim}: \{A(x) \mid x\in U\} \to [0, 1]$ confirms a fuzzy set $A^{\sim}$ on $U$, $A^{\sim}(x) = f^{\sim}(A(x))$. Call $A^{\sim}$ the intermediary negation set of $A$. Where

$$A^{\sim}(x) = \begin{cases} \lambda - \frac{2\lambda-1}{1-\lambda}(A(x)-\lambda), & \text{when } \lambda\in[½, 1) \text{ and } A(x)\in(\lambda, 1] \quad \text{(a)} \\ \lambda - \frac{2\lambda-1}{1-\lambda}A(x), & \text{when } \lambda\in[½, 1) \text{ and } A(x)\in[0, 1-\lambda) \quad \text{(b)} \\ 1-\frac{1-2\lambda}{\lambda}A(x)-\lambda, & \text{when } \lambda\in(0, ½] \text{ and } A(x)\in[0, \lambda) \quad \text{(c)} \\ 1-\frac{1-2\lambda}{\lambda}(A(x)+\lambda-1)-\lambda, & \text{when } \lambda\in(0, ½] \text{ and } A(x)\in(1-\lambda, 1] \quad \text{(d)} \\ A(x), & \text{other} \quad \text{(e)} \end{cases}$$

(*iii*). Mapping $f^{\neg}: \{A(x) \mid x\in U\} \to [0, 1]$ confirms a fuzzy set $A^{\neg}$ on $U$, $A^{\neg}(x) = f^{\neg}(A(x)) = max(A^{╕}(x), A^{\sim}(x))$. Call $A^{\neg}$ the contradictory negation set of $A$.

(2) If $A$ is a clear set, then $A(x)\in\{0, 1\}$, $A^{╕}(x) = 1-A(x)$, $A^{\sim}(x) = ½$, $A^{\neg}(x) = max(A^{╕}(x), A^{\sim}(x))$.

The set on the domain $U$ determined above is called "Sets with contradictory negation, opposite negation and intermediary negation", for short SCOI.

## 3.2 Logic with contradictory negation, opposite negation and intermediary negation

Based on SCOI, we further propose a "logic LCOI+PLCOI with contradictory negation, contrary negation, and intermediary negation" [8,9]. LCOI+PLCOI is a formal logical calculus system that extends the syntax and semantics of classical logic. LCOI is propositional logic, and PLCOI is predicate logic.

### 3.2.1 Definition of the logic LCOI+PLCOI

Symbols ¬, ╕, and ~ denote "contradictory negation", "opposite negation" and "intermediary negation" respectively. Symbols ∨, ∧ and → denote 'disjunction', 'conjunction' and 'implication', respectively. Symbol "├" denote formal deduction. The definition of logic LCOI+PLCOI is as follows

**Definition 1**. Let $\Im$ be set of atomic proposition. $\forall A\in\Im$, A is called well-formed formula (or formula). If A, B are formulas, then ¬A, ╕A, ~A, A→B, A∨B and A∧B are formulas.

(I) The following formulas as axioms:

(a1) $A\to(B\to A)$
(a2) $(A\to(A\to B))\to(A\to B)$
(a3) $(A\to B)\to((B\to C)\to(A\to C))$
(a4) $(A\to \neg B)\to(B\to \neg A)$
(a5) $(A\to \urcorner B)\to(B\to \urcorner A)$
(a6) $\neg A\to(A\to B)$
(a7) $((A \to \neg A)\to B)\to((A\to B)\to B)$
(a8) $A \to A \vee B$
(a9) $B \to A \vee B$
(a10) $A \wedge B \to A$
(a11) $A \wedge B\to B$
(a12) $\urcorner A \to \neg A \wedge \neg \sim A,\ \neg A \wedge \neg \sim A \to \urcorner A$
(a13) $\sim A \to \neg A \wedge \neg \urcorner A,\ \neg A \wedge \neg \urcorner A \to \sim A$

(II) The deduction rules:

[D1] $A_1, A_2, \ldots, A_n \vdash A_i\ (1\leq i \leq n)$
[D2] $A\to B, A \vdash B$

The logic calculus formal system determined above is called "propositional logic with contradictory negation, opposite negation and intermediary negation", for short LCOI.

On the basis of LCOI, adding predicates, individual words, quantifiers $\forall$ and $\exists$, as well as the following axioms and deduction rule, we can constitute a predicate logic PLCOI with contradictory negation, opposite negation and intermediary negation.

(I) Axioms:

(a14) $\forall x A(x) \to A(a)$
(a15) $A(a)\to\exists x A(x)$
(a16) $\forall x(A(x)\to B)\to\exists x(A(x)\to B)$
(a17) $\urcorner \forall x A(x) \to \exists x \urcorner A(x),\ \exists x \urcorner A(x) \to \urcorner \forall x A(x)$
(a18) $\urcorner \exists x A(x) \to\forall x \urcorner A(x),\ \forall x \urcorner A(x) \to \urcorner \exists x A(x)$

(II) Deduction rule:

[D3] If $\Sigma \vdash A(a)$ ($\Sigma$ is the set of formulas), where the individual constant $a$ does not appear in $\Sigma$, then $\Sigma \vdash\forall x A(x)$.

The logic calculus formal system determined above is called "predicate logic with contradictory negation, opposite negation and intermediary negation", for short PLCOI.

The propositional logic LCOI and predicate logic PLCOI are denoted as LCOI+PLCOI.

### 3.2.2 A continuous-valued semantics of LCOI+PLCOI

Regarding the semantics of the logic LCOI+PLCOI, we previously provided a three-valued semantics and proved the soundness theorem, completeness theorem and compactness theorem for LCOI+PLCOI under this semantics [8,9]. In order to make LCOI+PLCOI applicable in practice, we hereby propose continuous-valued semantics for LCOI+PLCOI with a truth domain of [0, 1].

Let $\Sigma$ be a set of formulas in LCOI+PLCOI, and let A be a formula in LCOI+PLCOI. Since LCOI+PLCOI is a formal logical system, defining the formal deduction $\Sigma \vdash A$ is provable in LCOI+PLCOI, just as it is in other formal logic.

**Definition 1**. The formal deduction $\Sigma \vdash A$ ($\Sigma$ can be empty set) is provable in LCOI+PLCOI, if there exists a finite sequence of formulas $E_1, E_2, \ldots, E_n$ such that $E_n = A$ and for each $E_n$ ($1 \leq k \leq n$), either $E_k$ is an axiom in LCOI+PLCOI or $E_k$ follows from $E_i$ and $E_j$ ($i < k, j < k$) using the deduction rule in LCOI+PLCOI, then $E_1, E_2, \ldots, E_n$ is called a "*proof*" of $\Sigma \vdash A$, n *length* of *proof*. $\Sigma \vdash A$ is denoted $\vdash A$ when $\Sigma$ is empty.

**Definition 2** (continuous-valued interpretation). Let $\Im$ be set of all formulas in LCOI+PLCOI, $\lambda\in(0, 1)$. $\forall A\in\Im$, mapping $\partial: \Im \rightarrow [0, 1]$ is called a $\lambda$-assignment of $\Im$, consists of the individual domain D and the following assignments for each constant symbol, function symbol and predicate symbol in A:

(1) for each constant symbol, assign an object in $D$ to correspond to it;

(2) for each n-variant function symbol, assign a mapping from $D^n$ to $D$ *to* correspond to it;

(3) for each n-variant predicate symbol, assign a mapping from $D^n$ to [0, 1] to correspond to it, and

[1] If A is an atomic formula, $\partial(A)$ takes only one value from [0, 1];

[2] $\partial(A)+\partial(\text{╕}A) = 1$;

[3]
$$\partial(\sim A) = \begin{cases} \lambda-\dfrac{2\lambda-1}{1-\lambda}(\partial(A)-\lambda), & \text{when } \lambda\in[½, 1) \text{ and } \partial(A)\in(\lambda, 1] \quad \text{(a)} \\ \lambda-\dfrac{2\lambda-1}{1-\lambda}\partial(A), & \text{when } \lambda\in[½, 1) \text{ and } \partial(A)\in[0, 1-\lambda) \quad \text{(b)} \\ 1-\dfrac{1-2\lambda}{\lambda}\partial(A)-\lambda, & \text{when } \lambda\in(0, ½] \text{ and } \partial(A)\in[0, \lambda) \quad \text{(c)} \\ 1-\dfrac{1-2\lambda}{\lambda}(\partial(A)+\lambda-1)-\lambda, & \text{when } \lambda\in(0, ½] \text{ and } \partial(A)\in(1-\lambda, 1] \quad \text{(d)} \\ \partial(A), & \text{other} \quad \text{(e)} \end{cases}$$

[4] $\partial(\neg A) = \max(\partial(\text{╕}A), \partial(\sim A))$;

[5] $\partial(A\vee B) = \max(\partial(A), \partial(B))$; $\partial(A\wedge B) = \min(\partial(A), \partial(B))$;

[6] $\partial(A\rightarrow B) = \Re(\partial(A), \partial(B))$. $\Re: [0, 1]^2 \rightarrow [0, 1]$ is a binary function;

[7] $\partial(\forall xP(x)) = \min\limits_{x\in D}\{\partial(P(x))\}$; $\partial(\exists xP(x)) = \max\limits_{x\in D}\{\partial(P(x))\}$.

In Definition 2, how to determine the truth value $\partial(\sim A)$ of the intermediary negation ~A of formula A in [0, 1] (i.e., the expression (a)-(e)), the parameter variable $\lambda$ ($\lambda\in(0, 1)$) is the key to the definition. The basic idea is as follows:

Since the truth values $\partial(A)$, $\partial(\neg A)$, $\partial(\text{╕}A)$, $\partial(\sim A)\in[0, 1]$, in order to determine their value range in [0, 1], we introduce a parameter variable $\lambda\in(0, 1)$. Consequently, when $\lambda \geq ½$, [0, 1] is divided into three sub-intervals: $[0, 1-\lambda)$, $[1-\lambda, \lambda]$, $(\lambda, 1]$. If $\partial(A)\in(\lambda, 1]$, then based on [2] in the definition, $\partial(\text{╕}A)\in[0, 1-\lambda)$. At this point, if $\partial(\sim A)\in[1-\lambda, \lambda]$, since $(\lambda, 1]$ and $[1-\lambda, \lambda]$ are disjoint intervals, then according to the principle that points in pairwise disjoint intervals in real variable functions have a one-to-one correspondence, the values in $(\lambda, 1]$ correspond one-to-one with those in$[1-\lambda, \lambda]$, and thus the expression (a) can be obtained. If $\partial(A)\in[0, 1-\lambda)$ and $\partial(\sim A)\in[1-\lambda, \lambda]$, we can similarly obtain expression (b). When $\lambda \leq ½$, [0, 1] is divided into three sub-intervals: $[0, \lambda)$, $[\lambda, 1-\lambda]$, $(1-\lambda, 1]$. In the same manner, we can establish expressions (c) and (d). For other scenarios, $\partial(\sim A) = \partial(A)$, which is expression (e).

For cases (a)–(d) in Definition 2, we illustrate them intuitively with the following figure (Figure 7). The symbols "•" and "o" in the figure represent the close endpoint and the open endpoint of an interval, respectively.

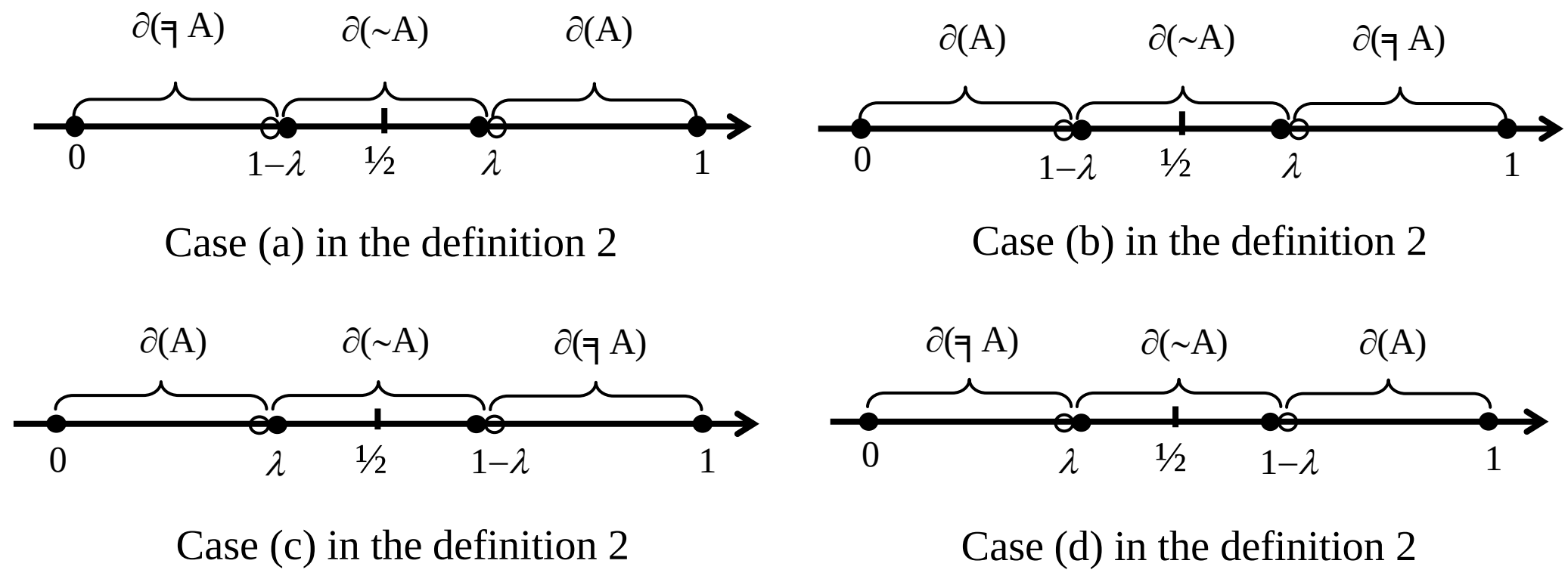

Case (a) in the definition 2

Case (b) in the definition 2

Case (c) in the definition 2

Case (d) in the definition 2

**Fig. 7** The interrelationships between $\partial(A)$, $\partial(\text{╕} A)$ and $\partial(\sim A)$

From the figure 7, it can be observed that:

(1) $\partial(\sim A)$ serves as an "intermediary" between $\partial(A)$ and its opposite $\partial(\text{╕} A)$, reflecting the important philosophy idea that "all opposing concepts transition to each other through an intermediary between them" [12].

(2) $\lambda\in(0, 1)$ is a variable parameter, it determines the value range of the membership degrees $\partial(A)$, $\partial(\text{╕} A)$ and $\partial(\sim A)$. That is, $\lambda$ is a "threshold" for the value range of these truth values. Its role and significance in practical applications will be discussed in detail in another article (a study on negation detection and negation resolution in medical texts).

As with the operational properties of SCOI [12, 13], it is easy to prove from Definition 2 that the truth values $\partial(A)$, $\partial(\neg A)$, $\partial(\text{╕} A)$ and $\partial(\sim A)$ have the following relationships and properties.

**Proposition 1**. If $\partial(A) = ½$, then

$$\partial(\neg A) = \partial(\text{╕} A) = \partial(\sim A) = ½.$$

**Proposition 2**. If $\lambda \geq ½$, then

$$\partial(A) > \partial(\sim A) > \partial(\text{╕} A) \text{ and } \partial(\neg A) = \partial(\sim A), \text{ if and only if } \partial(A)\in(\lambda, 1].$$

$$\partial(\text{╕} A) > \partial(\sim A) > \partial(A) \text{ and } \partial(\neg A) = \partial(\text{╕} A), \text{ if and only if } \partial(A)\in[0, 1-\lambda).$$

**Proposition 3**. If $\lambda \leq ½$, then

$$\partial(A) > \partial(\sim A) > \partial(\text{╕} A) \text{ and } \partial(\neg A) = \partial(\sim A), \text{ if and only if } \partial(A)\in(1-\lambda, 1].$$

$$\partial(\text{╕} A) > \partial(\sim A) > \partial(A) \text{ and } \partial(\neg A) = \partial(\text{╕} A), \text{ if and only if } \partial(A)\in[0, \lambda).$$

**Proposition 4**.

$$\partial(\sim A)\in[1-\lambda, \lambda], \text{ when } \lambda \geq ½.$$

$$\partial(\sim A)\in[\lambda, 1-\lambda], \text{ when } \lambda \leq ½.$$

The above propositions conveys the following practical significance: for a formula (proposition) A in the logic system LCOI+PLCOI and its contradictory negation ($\neg A$), opposite negation ($\text{╕} A$) and intermediary negation ($\sim A$), their truth values $\partial(A)$, $\partial(\neg A)$, $\partial(\text{╕} A)$ and $\partial(\sim A)$ (values within [0, 1]) relate to each other under different threshold values of $\lambda$.

In the semantics of mathematical logic, the metatheorems (such as the soundness theorem and completeness theorem) revolve around tautologies. "Tautology" (or logical truth) is a core concept. In classical two-valued logic, the truth value of a formula is either 0 or 1. The formula A is a tautology if and only if its truth value is always 1. In non-classical logics like fuzzy logic, since truth values include multiple values such as 0 and 1, the concept of "tautology' is weakened or modified (meaning A's truth value is not required to always be 1 but is required not to be lower than a certain value within the interval [0, 1]). This reflects different logical systems' interpretations of "logical truth". Thus, for the logic LCOI+PLCOI, which studies crisp and fuzzy entities, we provide a definition of "$\lambda$-tautology" based on Definition 2.

**Definition 3** ($\lambda$-tautology). Let $\Gamma$ be set of $\lambda$-assignment of $\Im$, $\forall A\in\Im$. For any $\lambda$-assignment $\partial \in \Gamma$, if $\partial(A) = 1$ then A is called a tautology. If $\partial(A) \geq \lambda$ ($\lambda > ½$), A is called a $\lambda$-tautology, and denoted $\models$ A. If there exists a $\lambda$-assignment $\partial \in \Gamma$ such that $\partial(A) \geq \lambda$, then A is called $\lambda$-satisfiable.

As in the proof methods of mathematical logic, the axioms in a logical system must be proven to be tautologies in order to prove the metatheorems of the logical system, such as the soundness theorem and completeness theorem. Therefore, all axioms in LCOI+PLCOI should be $\lambda$-tautologies. To this end, we need to determine the binary function $\Re$ in the definition 2.

**Definition 4**. Let $a, b\in[0, 1]$. The mapping $\Re^o$: $[0, 1]^2 \rightarrow [0, 1]$ is $\Re$, if satisfies:

$\Re^o(a, b) = 1$, when $a \leq b$. (1)

$\Re^o(a, b) = \max(1-a, b)$, when $a > b$. (2)

It can be easily proven that $\Re^o$ has the following properties.

**Proposition 5**. Let a, b$\in$[0, 1]. Then

a ≥ b, if and only if $\Re^o(a, c) \le \Re^o(b, c)$， for all c$\in$[0, 1]。 (3)

a > b, if and only if $\Re^o(c, a) \ge \Re^o(c, b)$。 for all c$\in$[0, 1]。 (4)

**Lemma 1**. For LCOI+PLCOI, if A and A→B are $\lambda$-tautologies, then B is a $\lambda$-tautology.

Proof: Let $\Re = \Re^o$, A and A→B are $\lambda$-tautologies. Suppose B is not a $\lambda$-tautology. Then, according to Definition 2, there exists a $\lambda$-assignment $\beta \in \Gamma$, $\beta(B) < \lambda$. Since A and A→B are $\lambda$-tautologies, so there are $\beta(A) \ge \lambda$ and $\beta(A \to B) \ge \lambda$. By [6] in the definition, $\beta(A \to B) = \Re^o(\beta(A), \beta(B)) \ge \lambda$. Due to $\beta(A) \ge \lambda$ and $\beta(B) < \lambda$，所以 $\beta(A) > \beta(B)$. Because $\lambda > ½$, so $\Re^o(\beta(A), \beta(B)) = \max(1-\beta(A), \beta(B)) < \lambda$ by (2). That is, it contradicts $\Re^o(\beta(A), \beta(B)) \ge \lambda$. Therefore, B is a $\lambda$-tautology. □

Based on Definition 2, let a, b and c represent $\partial(A)$, $\partial(B)$ and $\partial(C)$ respectively. If $\Re = \Re^o$, we can prove the following conclusion.

**Lemma 2**. Each axiom in LCOI+PLCOI is a $\lambda$-tautology.

Proof: Let $\Re = \Re^o$. According to [6] in Definition 2, the axioms (a1), (a2), and (a3) in LCOI+PLCOI can be expressed as follows:

(a1): $\Re(a, \Re(b, a)) \ge \lambda$, $(\lambda > ½)$

(a2): $\Re(\Re(a, \Re(a, b)), \Re(a, b)) \ge \lambda$, $(\lambda > ½)$

(a3): $\Re(\Re(a, b), \Re(\Re(b, c), \Re(a, c))) \ge \lambda$, $(\lambda > ½)$

For (a1). (i) If a ≤ b, then $\Re(a, \Re(b, a)) = \Re(a, \max(1-b, a))$ by the definition 4, where if 1−b > a, then $\Re(a, \max(1-b, a)) = \Re(a, 1-b) = 1 \ge \lambda$; if 1−b ≤ a, then $\Re(a, \max(1-b, a)) = \Re(a, a) = 1 \ge \lambda$. (ii) If a > b, then $\Re(b, a) \ge \Re(b, b)$ according to (4), $\Re(b, a) = 1$ by (1). So $\Re(a, \Re(b, a)) = \Re(a, 1)$. $\Re(a, \Re(b, a)) = 1$ by (1), i.e. $\Re(a, \Re(b, a)) \ge \lambda$. Therefore, the axiom (a1) is a $\lambda$-tautology by (i) and (ii).

For (a2). (i) If a ≤ b, then $\Re(a, \Re(a, b)) \ge \Re(b, \Re(a, b))$ by (3). According to (1), $\Re(a, b) = 1$. Hence, $\Re(\Re(a, \Re(a, b)), \Re(a, b)) = \Re(\Re(a, 1), 1)$. $\Re(\Re(a, 1), 1) = \Re(1, 1) = 1$ by (1), i.e. $\Re(\Re(a, \Re(a, b)), \Re(a, b)) \ge \lambda$. (ii) If a > b, based on the above proof, it is only necessary to prove $\Re(a, \Re(a, b)) \le \Re(a, b)$.

Suppose $\Re(a, \Re(a, b)) > \Re(a, b)$. $\Re(a, b) > b$ according to (4). $\Re(a, b) = \max(1-a, b) > b$ by (2). Hence, $\Re(a, b) = 1-a$. Substituting the hypothesis, there is $\Re(a, 1-a) > 1-a$, where if a ≤ 1−a, then $\Re(a, 1-a) = 1$ by (1); if a > 1−a, then $\Re(a, 1-a) = \max(1-a, 1-a) = 1-a$. $\Re(a, 1-a) = 1-a$ contradicts $\Re(a, 1-a) > 1-a$. Thus, $\Re(a, \Re(a, b)) \le \Re(a, b)$ holds true.

Therefore, the axiom (a2) is a $\lambda$-tautology by (i) and (ii).

For (a3). According to (1), only necessary to prove $\Re(a, b) \le \Re(\Re(b, c), \Re(a, c))$. Suppose $\Re(a, b) > \Re(\Re(b, c), \Re(a, c))$. From this, $\Re(\Re(b, c), \Re(a, c)) \ne 1$. According to (1), so

$$\Re(b, c) > \Re(a, c) \quad (5)$$

Thus, $\Re(a, c) \ne 1$. a > b by (3), a > c by (1). According to (2), $\Re(a, c) = \max(1-a, c)$. Substituting (5), $\Re(b, c) > \max(1-a, c)$, i.e. b ≤ c or b > c, there is $\Re(b, c) > \max(1-a, c)$.

If b ≤ c, $\Re(b, c) = 1$ according to (1). Hence, $\Re(b, c) > \max(1-a, c)$.

If b > c, $\Re(b, c) = \max(1-b, c)$ according to (2). Because of a > b, i.e. 1−a < 1−b, Hence, $\Re(b, c) = \max(1-b, c) > \max(1-a, c)$. However, if 1−b ≤ c, then $\max(1-b, c) = c$ and $\max(1-a, c) = c$, it contradicts $\max(1-b, c) > \max(1-a, c)$. So, only when b > c and 1−b > c, $\Re(b, c) > \max(1-a, c)$.

Substitute a > b, b > c, 1−b > c into the suppose: $\Re(a, b) > \Re(\Re(b, c), \Re(a, c))$, then $\max(1-a, b) > \max(1-\max(1-b, c), \max(1-a, c)) = \max(b, \max(1-a, c))$ by (3). However, when 1−a < b, there is $\max(1-a, b) = b > \max(b, \max(1-a, c)) = b$, the two are contradictory; when 1−a ≥ b, there is $\max(1-a, b) = 1-a > \max(b, \max(1-a, c)) = 1-a$, the two are contradictory. Thus, the assumption $\Re(a, b) > \Re(\Re(b, c), \Re(a, c))$ is not valid. Therefore, the axiom (a3) is a $\lambda$-tautology.

Similarly, it can be proven that if $\Re = \Re^o$, then the axioms (a4) – (a18) in LCOI+PLCOI are all $\lambda$-tautologies. □

Based on the above results, similar to the proof methods (method of induction) for soundness in mathematical logic, we can prove the following soundness theorem for LCOI+PLCOI.

**Theorem 1** (*Soundness theorem*). Let $\Phi$ ($\Phi \subseteq \Im$) be a set of formulas in LCOI+PLCOI and A be a formula in LCOI+PLCOI.

(a) If $\vdash$ A, then $\models$ A.

(b) If $\Phi \vdash$ A, then $\Phi \models$ A.

Proof: If $\vdash$ A, then A is provable in LCOI+PLCOI. According to the definition 1, Induct on the length $n$ of the sequence of formulas $A_1, A_2, \ldots, A_n$ for the proof of A.

(i) When $n = 1$, then according to Definition 1 in Section 5.3, $A_1$ (which is A) is an axiom in LCOI+PLCOI. By Lemma 2, A is a $\lambda$-tautology.

(ii) Suppose the theorem holds for $k < n$. That is, all proof sequences of A with fewer than n steps are $\lambda$-tautologies. We prove that the theorem holds when $n = k$. According to Definition 1 in Section 5.3, there are two cases: (1) $A_n$ is an axiom of LCOI+PLCOI, or (2) $A_n$ is a formula deduced from $A_i$ and $A_i$ and $A_j$ ($i < n, j < n$) using the deduction rule [D2] in LCOI+PLCOI. If it is case (1), then it is similar to the proof of (i). If it is case (2), the formal expressions of the formulas $A_i$ and $A_j$ must be B and B→A. According to the hypothesis, B and B→A are $\lambda$-tautologies. Therefore, by Lemma 1, A (i.e. $A_n$) is a $\lambda$-tautology. According to the principle of mathematical induction and Definition 3, $\models$ A holds.

When $\Phi$ is the empty set, (b) is equivalent to (a). Therefore, (b) can be proven similarly. □

We need to point out that when $\Re = \Re^o$, it can be verified that the completeness theorem for LCOI+PLCOI does not hold. Whether there exists a specific $\Re$ such that the completeness theorem for LCOI+PLCOI holds will be discussed in another article.

## 4. FCACOI: formal concept analysis with three types of negation

In classical Formal Concept Analysis (FCA), a formal context (a two-dimensional table) records “which objects have which attributes”, from which knowledge units (formal concepts) are derived. Each formal concept consists of a set of objects (extent) and the set of attributes they commonly share (intent). All these concepts are arranged in a hierarchical structure (concept lattice) ordered by extent from small to large (and intent from large to small) [6, 13].

In extended FCA research, the classical binary logic’s limitations in handling uncertainty, incomplete information and dynamic data are often addressed by combining FCA with different theories [7, 13]. For example, “Fuzzy Formal Concept Analysis” employs fuzzy sets to characterize the binary fuzzy relations between object sets and attribute sets, thus extending FCA [14]; “Three-Way Formal Concept Analysis” introduces “positive and negative operators”, allowing concepts to express both “commonly possessed” and “commonly not possessed” information simultaneously [15]; “Probabilistic Formal Concept Analysis” introduces probabilistic measures to handle the randomness in the relationships between objects and attributes [16]; and so forth.

FCA primarily focuses on positive relationships between objects and attributes. Neither its original definition nor its standard form involves negation; it includes only affirmative information and contains no negation. In other words, FCA lacks the mechanism to handle negation and its relations. On the other hand, for FCA and its aforementioned extensions, binary logic or non-classical logics (such as many-valued logic, fuzzy logic, probabilistic logic, etc.) serve as their logical foundations. Since these logical theories’ basic concepts and formal languages include only one kind of negation (i.e., classical negation), FCA and its extensions cannot distinguish or express the three types of negation present in information and knowledge: contradictory negation, contrary negation, and intermediary negation.

To overcome this limitation, in this section, we introduce the concepts of contradictory negation, contrary negation and intermediary negation into FCA. Based on the set SCOI and the logic LCOI+PLCOI, we provide a formal context, Galois connection operators, formal concept and concept lattice with three types of negation. We propose a “Formal Concept Analysis with contradictory negation, opposite negation and intermediary negation”, abbreviated as FCACOI. For the first time within the FCA framework, explicit representation and reasoning about these three negations and their relationships are realized, thereby expanding and deepening FCA and enhancing its capacity to reveal and represent complex relationships within knowledge and data.

### 4.1 A formal context with three types of negation and its consistency constraint

A formal context (FC) is a core concept in Formal Concept Analysis theory. It provides the foundation for the concept lattice by establishing the relationship between objects and attributes. A classical formal context is typically defined as a triple $(U, A, I)$, where $U$ is the set of objects, $A$ is the set of attributes and $I$: $U\times A \rightarrow \{0, 1\}$ is a binary relation between $U$ and $A$, with $I \subseteq U\times A$. In a classical formal context, if an object $x\in U$ has an attribute $a\in A$, then $I(x, a) = 1$. Otherwise, object $x$ does not have attribute $a$, i.e., $I(x, a) = 0$. A fuzzy formal context (FFC) is an extension of the classical formal context. It provides the foundation for the fuzzy concept lattice by establishing fuzzy or uncertain relationships between objects and attributes. In a fuzzy formal context $(U, A, I)$, the relation $I$: $U\times A \rightarrow [0, 1]$ is a binary fuzzy relation between $U$ and $A$, representing the membership degree of object $x\in U$ to attribute $a\in A$ (i.e., the degree to which object $x$ has attribute $a$) [14, 17].

The intuitive meaning of a triple is the relationship established between the subject and the object as defined by the property, and the uncertainty within a triple is inevitably reflected by the uncertainty that exists in the relationship between subject and object [18]. Similarly, the intuitive meaning of a formal context is determined by the relation between objects and attributes, and the uncertainty of the formal context must be reflected through the uncertainty in the relation between objects and attributes. Therefore, establishing the relationship between objects and attributes is crucial in various formal contexts.

For a formal context, we consider that the relationship between objects and attributes should be distinguished as follows:

- For an object $x\in U$, if attribute $a\in A$ is a clear attribute, then the relation $I$ between $x$ and $a$ is a binary clear relation: a composite mapping from $U$ to $A$ and then to $\{0, 1\}$.
- For an object $x\in U$, if attribute $a\in A$ is a fuzzy attribute, then the relation $I$ between $x$ and $a$ is a binary fuzzy relation: a composite mapping from $U$ to $A$ and then to $[0, 1]$.

The above two binary relations can be mathematically expressed as follows.

(1) For an object $x\in U$, if exists an attribute $a\in A$ and $a$ is clear attribute, then relation $I$ between $x$ and $a$ is a composite mapping from $U$ to $A$ and then to $\{0, 1\}$, i.e., $I = f_2 \circ f_1$. Here $f_1 : U \rightarrow A$, $f_2 : A \rightarrow \{0, 1\}$, $I \subseteq U\times A$. $I = \{(x, f_2(a)) \mid x\in U, f_2(a)\in\{0, 1\}\}$, $I(x, a)\in\{0, 1\}$.

(2) For an object $x\in U$, if exists an attribute $a\in A$ and $a$ is fuzzy attribute, then relation $I$ between $x$ and $a$ is a composite mapping from $U$ to $A$ and then to $[0, 1]$, i.e., $I = f_2 \circ f_1$. Here $f_1 : U \rightarrow A$, $f_2 : A \rightarrow \{0, 1\}$, $I \subseteq U\times A$. $I = \{(x, f_2(a)) \mid x\in U, f_2(a)\in[0, 1]\}$, $I(x, a)\in[0, 1]$.

It can be seen that the binary relations $I$ in (1) and (2) are respectively clear sets and fuzzy sets on $U\times A$. Since a clear set is a special case of a fuzzy set (a fuzzy set extends the range of the characteristic function (membership function) of a classical set from $\{0, 1\}$ to $[0, 1]$), it is thus conventionally assumed that $I(x, a)\in[0, 1]$. Here, $I(x, a)$ represents the degree to which an object $x\in U$ possesses an attribute $a\in A$; $I(x, a) = 1$ means that object $x$ possesses attribute $a$, and $I(x, a) = 0$ means that object $x$ does not possess attribute $a$. Therefore, $I$ collectively express the binary relation between the object set $U$ and the attribute set $A$ in both the classical formal context FC and the fuzzy formal context FFC.

Based on the above, we take the set SCOI and the logic LCOI+PLCOI as the foundation, distinguish three types of negation for clear and fuzzy attributes: contradictory negation, opposite negation and intermediary negation, and propose a “formal context FCCOI with contradictory negation, opposite negation and intermediary negation” as follows.

**Definition 1** (Formal Context FCCOI). Let $U$ be a non-empty finite set of objects, $A_{total} = A \cup A^{\neg} \cup A^{\urcorner} \cup A^{\sim}$ be the set of attributes, and $I$: $U \times A_{total} \rightarrow [0, 1]$ is a binary relation between $U$ and $A_{total}$, $I \subseteq U \times A_{total}$. Then the triple

$$(U, A_{total}, I)$$

is called a formal context with contradictory negation, opposite negation and intermediary negation, denoted as FCCOI. Here, $A = \{ a_1, a_2, ..., a_n\}$ represents the original attribute set of objects in $U$; $A^{\neg} = \{a_i^{\neg} \mid a_i \in A\}$, $A^{\urcorner} = \{a_i^{\urcorner} \mid a_i \in A\}$ and $A^{\sim} = \{a_i^{\sim} \mid a_i \in A\}$ denote the sets of contradictory negation, opposite negation and intermediary negation of $A$, respectively; $a_i^{\neg}$, $a_i^{\urcorner}$ and $a_i^{\sim}$ denote the contradictory negation, opposite negation and intermediary negation of attribute $a_i$, respectively.

From Definition 1, although FCCOI differs structurally from the classical formal context FC and the fuzzy formal context FFC, it reflects the three types of negations on attributes and their relationships while preserving the fundamental meanings of FC and FFC. In other words, the classical formal context FC and fuzzy formal context FFC can be viewed as degenerate cases of FCCOI when only one type of negation (i.e., contradictory negation) is employed.

In Definition 1, the relation $I \subseteq U \times A_{total}$ records whether objects in $U$ "possess" the original attributes and their three types of negated attributes, as well as the degree to which they possess them. Concretely, $I(x, a_i)$, $I(x, a_i^{\neg})$, $I(x, a_i^{\urcorner})$, $I(x, a_i^{\sim}) \in [0, 1]$ represent the degrees to which object $x \in U$ possesses attributes $a_i$, $a_i^{\neg}$, $a_i^{\urcorner}$ and $a_i^{\sim}$, respectively.

The set SCOI and the continuous-valued semantics of the logic LCOI+PLCOI lay the foundation for the semantics of the formal context FCCOI. For $I(x, a_i)$, $I(x, a_i^{\neg})$, $I(x, a_i^{\urcorner})$ and $I(x, a_i^{\sim})$ in FCCOI, we can replace the mapping $f$ in the dfinition of set SCOI (Definition 1 in Section 3.1) with $I$. Then, the degree to which the object $x$ in FCCOI has the attribute $a_i$, namely $I(x, a_i)$, corresponds to the membership degree $A(x)$ of element $x$ to the set $A$ in SCOI; $I(x, a_i^{\neg})$, $I(x, a_i^{\urcorner})$ and $I(x, a_i^{\sim})$ correspond respectively to the membership degrees $A^{\neg}(x)$, $A^{\urcorner}(x)$, $A^{\sim}(x)$ of $x$ to the sets $A^{\neg}$, $A^{\urcorner}$ and $A^{\sim}$ in SCOI. Therefore, $I(x, a_i)$, $I(x, a_i^{\neg})$, $I(x, a_i^{\urcorner})$ and $I(x, a_i^{\sim})$ can be computed through the membership degrees $A^{\neg}(x)$, $A^{\urcorner}(x)$, $A^{\sim}(x)$ of sets in SCOI.

In like manner, by replacing the mapping $\partial$ in the continuous-valued interpretation of the logic LCOI+PLCOI (Definition 2 in Section 3.2.2) with $I$. Then, $I(x, a_i)$ can be interpreted as the truth value $\partial(\mathrm{A})$ of the formula $A$: "object $x$ has attribute $a_i$" in LCOI+PLCOI; $I(x, a_i^{\neg})$, $I(x, a_i^{\urcorner})$ and $I(x, a_i^{\sim})$ correspond respectively to the truth values $\partial(\neg\mathrm{A})$, $\partial(\urcorner \mathrm{A})$ and $\partial(\sim\mathrm{A})$ of the formula $\neg$A "object $x$ has the contradictory negation $a_i^{\neg}$ of attribute $a_i$", the formula $\urcorner$A "object $x$ has the opposite negation $a_i^{\urcorner}$ of attribute $a_i$", and the formula $\sim$A "object $x$ has the intermediary negation $a_i^{\sim}$ of attribute $a_i$". Therefore, $I(x, a_i)$, $I(x, a_i^{\neg})$, $I(x, a_i^{\urcorner})$, $I(x, a_i^{\sim})$ can be determined by the continuous-valued interpretation $\partial$ of the logic LCOI+PLCOI.

**Example 1**. The following table (Table 1) describes a formal context FCCOI: $U = \{x_1, x_2, x_3, x_4\}$, $A_{total} = A \cup A^{\neg} \cup A^{\urcorner} \cup A^{\sim}$, with the original attribute set $A = \{ a_1, a_2, a_3\}$. Each value in the table represents the degree to which an object possesses a certain attribute. For example, $I(x_1, a_2^{\neg}) = 0.5$ indicates that the degree to which object $x_1$ possesses attribute $a_2^{\neg}$ is 0.5. All values are computed through the definition of the set SCOI (Definition 1 in Section 3.1). The specific computation process will be discussed in Section 5.2.

**Table 1**. Formal context FCCOI with three types of negation

| $A_{total}$ / $U$ | $A \cup A^{\neg} \cup A^{\urcorner} \cup A^{\sim}$ | | | | | | | | | | | |
|---|---|---|---|---|---|---|---|---|---|---|---|---|
| | $a_1$ | $a_1^{\neg}$ | $a_1^{\urcorner}$ | $a_1^{\sim}$ | $a_2$ | $a_2^{\neg}$ | $a_2^{\urcorner}$ | $a_2^{\sim}$ | $a_3$ | $a_3^{\neg}$ | $a_3^{\urcorner}$ | $a_3^{\sim}$ |
| $x_1$ | 0.9 | | | | | 0.5 | | | | 0.55 | | |
| $x_2$ | | | 0.1 | | | | 0.2 | | 0.7 | | | |
| $x_3$ | | | | 0.45 | | | 0.2 | | | | | 0.55 |
| $x_4$ | 0.9 | | | | 0.8 | | | | | 0.55 | | |

For a formal context, the consistency constraint is a necessary fundamental guarantee. The purpose of consistency constraint is to ensure that the formal context is logically contradiction-free and logically self-consistent (i.e., no contradictions can be derived within the logical theory) [19]. The classical formal context FC is based on binary logic, and its relation $I \subseteq U \times A$ inherently implies the most basic non-contradiction: an object-attribute pair either belongs to $I$ (has the attribute) or does not (does not have the attribute), with the law of excluded middle holding between the two. Therefore, this binary structure automatically ensures the non-contradiction and logical self-consistency of FC. As a result, FC usually does not require explicit statement of consistency constraints [19, 20].

For the formal context FCCOI with three types of negation, it is an extension of the classical formal context FC based on the set SCOI and the logic LCOI+PLCOI, where the attribute set is extended to the union of the original attribute set and the three types of negation sets. Hence, the consistency constraints aim to ensure that the introduction of the three types of negation attributes accurately corresponds to the semantics of the three negations in the set SCOI and the logic LCOI+PLCOI, without compromising the original non-contradiction and maintaining logical self-consistency (i.e., no contradictions can be derived in the logic LCOI+PLCOI). This lays a reliable foundation for the subsequent construction of concept lattices and logical reasoning.

Specifically, the consistency constraints and their meanings for the formal context $(U, A_{total}, I)$ with three types of negation are as follows:

- Mutual exclusion: For any $x \in U$ and $a \in A$, exactly one of the four holds: $(x, a) \in I$, $(x, a^{\neg}) \in I$, $(x, a^{\urcorner}) \in I$ and $(x, a^{\sim}) \in I$. That is, each object-attribute pair has only one binary relation $I$. The mutual exclusion requires that affirmation and any one of the three types of negation cannot simultaneously hold, ensuring no logical contradiction arises for an object-attribute pair.

- Involution: For any $x \in U$ and $a \in A$, if $(x, a^{\neg}) \in I$, then $(x, (a^{\neg})^{\neg}) = (x, a) \in I$; if $(x, a^{\urcorner}) \in I$, then $(x, (a^{\urcorner})^{\urcorner}) = (x, a) \in I$. That is, double contradictory negation and double opposite negation restore the original attribute. The logic LCOI+PLCOI provides the theoretical basis for this (see Theorem 5 in Section 5.2.1 of [9]). In LCOI+PLCOI, a formula A and its double opposite negation $\urcorner\urcorner A$ and double contradictory negation $\neg\neg A$ are mutually derivable, i.e., $A \vdash \urcorner\urcorner A$; $\urcorner\urcorner A \vdash A$; $A \vdash \neg\neg A$; $\neg\neg A \vdash A$. Involution ensures that the contradictory and contrary negation operators form an involution at the conceptual level, providing a basis for involution properties in subsequent lattice operations.

- Completeness: Ensure that for any object $x \in U$, the attribute it possesses can only be one of the four attributes $a$, $a^{\neg}$, $a^{\urcorner}$ and $a^{\sim}$. In other words, each object-attribute pair in FCCOI is in exactly one of the four states: $(x, a) \in I$, $(x, a^{\neg}) \in I$, $(x, a^{\urcorner}) \in I$ and $(x, a^{\sim}) \in I$. This constraint makes the formal context complete (i.e., no unknown states for object-attribute pairs), all information is explicitly recorded, thus preventing information loss when constructing concept lattices.

The above consistency constraints of the formal context FCCOI with three types of negation, together ensure that FCCOI is a well-formed structure semantically consistent with the set SCOI and the logic LCOI+PLCOI, thereby supporting the construction of formal concepts and concept lattices featuring three types of negation.

## 4.2 Formal concept and concept lattice with three types of negation

In Formal Concept Analysis, the "formal concept" defined on a formal context unifies the extension and intension of a concept, characterizing the formal concept by an "object set" and an "attribute set" (which are mutually closed under each other). This makes the concept a "complete" cognitive unit and naturally forms a hierarchical structure among concepts, thus providing a rigorous mathematical definition of concepts [5, 6].

A formal concept on a classical or fuzzy formal context $(U, A, I)$ is defined as an ordered pair $(X, B)$ where $X \subseteq U$, $B \subseteq A$. Given operators $f$ and $g$, if they satisfy $B = f(X) = \{a \in A \mid \forall x \in X, (x, a) \in I\}$, $X = g(B) = \{x \in U \mid \forall a \in B,$ $(x, a)\in I\}$, then $X$ is called the extension and $B$ the intension of the formal concept. Here, $f(X)$ represents the set of attributes common to all objects in $X$, and $g(B)$ represents the set of objects possessing all attributes in $B$. It is evident that the operators $f$ and $g$ are necessary conditions for the definition [5, 14].

For the formal concept on the formal context FCCOI, since FCCOI distinguishes three types of negations of the original attribute set $A$, a formal concept on FCCOI should be able to reflect $A$ and its three kinds of negation. To determine the formal concept on FCCOI, we define four Galois connection operators corresponding respectively to affirmation, contradictory negation, opposite negation, and intermediary negation. Each operator independently preserves the Galois connection property of classical FCA and establishes intrinsic logical connections through the set SCOI and the logic LCOI+PLCOI. These operators collectively support the definition of "formal concept with three types of negation", enabling the concept intension to simultaneously express four logical states. This lays a solid algebraic foundation for constructing concept lattices and reasoning frameworks capable of handling complex negation relations.

**Definition 2** (Galois connection operators). For a formal context FCCOI with three types of negation, let $X\subseteq U$, $A\subseteq A_{total}$ be set of original attributes. The following four operators on FCCOI are the Galois connection operators:

Affirmation operator $(\cdot)^+$: $X^+ = \{a\in A \mid \forall x\in X, (x, a)\in I\}$; $B^+ = \{x\in U \mid \forall a\in A, (x, a)\in I\}$

Contradictory negation operator $(\cdot)^c$: $X^c = \{a\in A \mid \forall x\in X, (x, a^{\neg})\in I\}$; $C^c = \{x\in U \mid \forall a\in A, (x, a^{\neg})\in I\}$

Opposite negation operator $(\cdot)^o$: $X^o = \{a\in A \mid \forall x\in X, (x, a^{╕})\in I\}$; $O^o = \{x\in U \mid \forall a\in A, (x, a^{╕})\in I\}$

Intermediary negation operator $(\cdot)^d$: $X^d = \{a\in A \mid \forall x\in X, (x, a^{\sim})\in I\}$; $D^d = \{x\in U \mid \forall a\in A, (x, a^{\sim})\in I\}$

The meanings and characteristics of the above four Galois connection operators are as follows:

1. The operator $(\cdot)^+$ is the same as the operators $f$ and $g$ in the classical definition of formal concept. $X^+$ denotes the set of attributes commonly possessed by the objects in $X$, and $B^+$ denotes the set of objects that have all the attributes in $A$.
2. In the operators $(\cdot)^c$, $(\cdot)^o$ and $(\cdot)^d$, $X^c$, $X^o$ and $X^d$ respectively denote the sets of attributes of the objects in $X$ that possess the contradictory negation $A^{\neg}$, the opposite negation $A^{╕}$, and the intermediary negation $A^{\sim}$ of the original attribute set A; $C^c$, $O^o$ and $D^d$ respectively denote the sets of objects that possess $A^{\neg}$, $A^{╕}$ and $A^{\sim}$.

As with the proof of the basic properties of the classical Galois connection [6], it can be demonstrated that all four Galois connection operators satisfy the fundamental properties of the classical Galois connection.

**Property 1**. Let $(U, A_{total}, I)$ be a formal context; $X, X_1, X_2 \subseteq U$; $B, B_1, B_2\subseteq A$ ($A$ is the set of original attributes, $A\subseteq A_{total}$). For any operator $*\in\{+, c, o, d\}$, the following hold:

(1) Fundamental lemma: $X^* = B \Leftrightarrow B^* = X$

(2) Antitonicity: $X_1 \subseteq X_2 \Rightarrow X_2^*\subseteq X_1^*$; $B_1 \subseteq B_2 \Rightarrow B_2^*\subseteq B_1^*$

(3) Extensivity: $X \subseteq X^{**}$; $B \subseteq B^{**}$

(4) Idempotency: $X^{***} = X^*$; $B^{***} = B^*$.

Based on the set SCOI and the logic LCOI+PLCOI [8, 9], it can be proven that the operators $(\cdot)^+$, $(\cdot)^c$, $(\cdot)^o$, $(\cdot)^d$ also have the following special properties.

**Property 2**. Let $(U, A_{total}, I)$ be a formal context. Then:

(1) Implication:

$X^c \supseteq X^o$, $C^c \supseteq O^o$ (contradictory negation implies opposite negation)

$X^c \supseteq X^d$, $C^c \supseteq D^d$ (contradictory negation implies intermediary negation)

(2) Mutual exclusivity:

$X^+\cap X^c = \varnothing$, $X^+\cap X^o = \varnothing$, $X^+\cap X^d = \varnothing$ (the attribute set $X^+$ is disjoint with its three kinds of negation)

$X^o \cap X^d = \varnothing$ (the opposite negation and intermediary negation of attribute set $X^+$ are disjoint)

(3) Completeness:

$A_{total} = X^+ \cup X^c \cup X^o \cup X^d$ (consistent with Definition 1)

These properties guarantee that each operator can independently define a Galois connection pair. These Galois connections are interlinked through consistency constraints of the formal context $(U, A_{total}, I)$ with three kinds of negation. Therefore, we can define the "formal concept with three types of negation" as follows.

**Definition 3** (Formal Concept with three types of negation). Let $B$ be the intension of a classical formal concept; $C$, $O$ and $D$ be the contradictory negation, opposite negation and intermediary negation of $B$, respectively. A formal concept with three types of negation (abbreviated as concept) on the formal context ($U$, $A_{total}$, $I$) is a quintuple:

$$(X, B, C, O, D)$$

where $X \subseteq U$; $B, C, O, D \subseteq A_{total}$. They satisfy the following closure conditions:

Affirmation closure: $X^+ = B$, $B^+ = X$,

Contradictory negation closure: $X^c = C$, $C^c = X$,

Opposite negation closure: $X^o = O$, $O^o = X$,

Intermediary negation closure: $X^d = D$, $D^d = X$,

$X$ is called the extension of the concept, and the ordered quadruple ($B$, $C$, $O$, $D$) is called the intension of the concept.

In classical FCA, the intension of a formal concept is an attribute set. Definition 3 indicates that, in a formal concept with three types of negation, the intension is no longer a single attribute set but an ordered quadruple ($B$, $C$, $O$, $D$) that includes the affirmative intension and its three kinds of negation. The four independent intension components $B$, $C$, $O$ and $D$, each have distinct logical meanings (affirmation, contradictory negation, opposite negation, intermediary negation), corresponding respectively to the original attribute set $A$, its contradictory negation set $A^{\neg}$, opposite negation set $A^{╕}$ and intermediary negation set $A^{\sim}$ within $A_{total}$.

For formal concept with three types of negation, the operations of inclusion, intersection, and union are completely consistent with those of classical formal concept. Moreover, the single closure (affirmative closure) of classical formal concept is extended to four closures that include itself and its three negations. Therefore, formal concepts with three types of negation expand the properties possessed by classical formal concepts. Based on the above, it can be proven that formal concepts with three types of negation have the following properties.

Let $X, X_1, X_2 \subseteq U$; $B, B_1, B_2 \subseteq A$ ($A$ is the set of original attributes, $A \subseteq A_{total}$). For any operator $* \in \{+, c, o, d\}$, the formal concept with three types of negation has the following properties:

(1) $X_1 \subseteq X_2 \Rightarrow X_2^* \subseteq X_1^*$, $B_1 \subseteq B_2 \Rightarrow B_2^* \subseteq B_1^*$

(2) $X \subseteq X^{**}$, $B \subseteq B^{**}$

(3) $X^{***} = X^*$, $B^{***} = B^*$

(4) $X \subseteq B^* \Leftrightarrow B \subseteq X^*$

(5) $(X_1 \cup X_2)^* = X_1^* \cap X_2^*$, $(B_1 \cup B_2)^* = B_1^* \cap B_2^*$

(6) $(X_1 \cap X_2)^* \supseteq X_1^* \cup X_2^*$, $(B_1 \cap B_2)^* \supseteq B_1^* \cup B_1^*$

(7) $(X^{**}, X^*)$ and $(B^{**}, B^{**})$ are concepts.

Based on the order relations between concepts on the classical formal context [6,7], we can define the order relation between formal concepts with three types of negation as follows.

**Definition 4** (partial order relation $\leq$). Let $T_1 = (X_1, B_1, C_1, O_1, D_1)$ and $T_2 = (X_2, B_2, C_2, O_2, D_2)$ be two formal concepts with three types of negation on the formal context ($U$, $A_{total}$, $I$). $T_1$ is less than or equal to $T_2$ (denoted as $T_1 \leq T_2$) , if and only if there is an inclusion relation between their extensions. That is,

$$T_1 \leq T_2, \text{ if and only if } X_1 \subseteq X_2.$$

According to (1) and (2) in Property 1 above, the above definition is equivalent to:

$$T_1 \leq T_2 \text{ if and only if } B_2 \subseteq B_1, C_2 \subseteq C_1, O_2 \subseteq O_1, D_2 \subseteq D_1.$$

The concept lattice, as a method for data analysis and knowledge modeling, structures data into a hierarchical knowledge representation by revealing the relationships between objects and attributes, thereby providing a solid mathematical foundation (a complete lattice) for knowledge representation [19]. Based on the above, we propose a "concept lattice with contradictory negation, opposite negation and intermediary negation" as follows.

**Definition 5** (concept lattice CLCOI). All formal concepts with three types of negation on a formal context ($U$,

$A_{total}$, $I$) form a complete lattice under the partial order $\leq$. The supremum and infimum are defined as follows:

Infimum: $T_1 \wedge T_2 = (X_1 \cap X_2, (B_1 \cup B_2)^{++}; (C_1 \cup C_2)^{cc}, (O_1 \cup O_2)^{oo}, (D_1 \cup D_2)^{dd})$.

Supremum: $T_1 \vee T_2 = ((X_1 \cup X_2)^{++}; (B_1 \cap B_2), (C_1 \cap C_2), (O_1 \cap O_2), (D_1 \cap D_2))$.

This complete lattice is called the concept lattice CLCOI with contradictory negation, opposite negation and intermediary negation, abbreviated as CLCOI.

*Remark* 1. The symbol "++" denotes applying the affirmative operator $(\cdot)^+$ twice (i.e., the affirmative closure). The symbols *cc*, *oo* and *dd* denote respectively applying the operators $(\cdot)^c$, $(\cdot)^o$ and $(\cdot)^d$ twice (i.e., the contradictory negation, opposite negation and intermediary negation closures). Applying the operators twice ensures that the result satisfies the closure conditions of the Galois connection, thereby yielding a new formal concept.

*Remark* 2. In the infimum, the extensions take the intersection (common objects), while the intensions take the union of each component followed by closure (because the union may not be closed, closure is needed to obtain the maximal common attribute set). In the supremum, the extensions take the union followed by closure (because the union may not be closed), while the intensions take the intersection (common attributes; intersections are always closed).

Regarding the relationship between the concept lattice CLCOI and the classical concept lattice CL and fuzzy concept lattice FCL, it is evident from the above that when ignoring the contradictory negation set $A^{\neg}$, the opposite negation set $A^{⌉}$ and the intermediary negation set $A^{\sim}$ within the extended attribute set $A_{total}$ (i.e., retaining only the original attribute set $A$), the concept lattice CLCOI degenerates into CL and FCL. This shows that the concept lattice CLCOI with three types of negation is an extension of the classical concept lattice CL and the fuzzy concept lattice FCL, and it maintains consistency with the notion of "partial order" in both CL and FCL.

The concept lattice CLCOI with three types of negation, not only preserves all the favorable algebraic properties of the classical concept lattice CL and the fuzzy concept lattice FCL, but also, due to the diversity of intension components, naturally reflects the hierarchy and dependencies among different negation information through the partial order relation in the lattice.

For a formal context, formal concept and concept lattice with three types of negation, we take as an example a formal context FCCOI (Table 1), all the formal concepts derived from it, as well as the concept lattice CLCOI formed by these formal concepts and its Hasse diagram.

**Example 2**. In a formal context FCCOI (Table 1), $U = \{x_1, x_2, x_3, x_4\}$, $A_{\text{total}} = A \cup A^{\neg} \cup A^{⌉} \cup A^{\sim}$, with the original attribute set $A = \{ a_1, a_2, a_3\}$. All concepts derived from Table 1 are shown in Table 2. Here, the value following an attribute indicates the degree to which an object possesses that attribute; for example, $a_1(0.9)$ means that the object $x_1$ possesses attribute $a_1$ to the degree of 0.9.

**Table 2**. All formal concepts derived from Table 1

| formal concept | (extension, intension) |
|---|---|
| $FC_1$ | $(\{x_1, x_2, x_3, x_4\}, \varnothing)$ |
| $FC_2$ | $(\{x_1, x_4\}, \{a_1(0.9), a_3^{\neg}(0.55)\}$ |
| $FC_3$ | $(\{x_2, x_3\}, \{a_2^{⌉}(0.2) \}$ |
| $FC_4$ | $(\{x_1\}, \{a_1(0.9), a_2^{\neg}(0.5), a_3^{\neg}(0{,}55)\})$ |
| $FC_5$ | $(\{x_2\}, \{a_1^{⌉}(0.1), a_2^{⌉}(0.2), a_3(0{,}7)\})$ |
| $FC_6$ | $(\{x_3\}, \{a_1^{\sim}(0.45), a_2^{⌉}(0.2), a_3^{\sim}(0{,}55)\})$ |
| $FC_7$ | $(\{x_4\}, \{a_1^{⌉}(0.1), a_2(0.8), a_3^{\neg}(0{,}55))$ |
| $FC_8$ | $(\varnothing,\{a_1(1), a_2(1), a_3(1)\})$ |

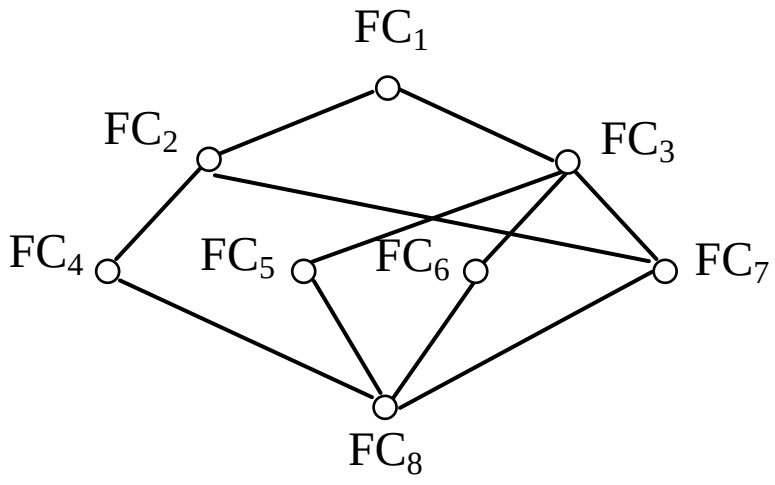

**Figure 8**. Hasse diagram of the concept lattice CLCOI

In summary, for the formal context, formal concept, concept lattice and formal concept analysis with three types of negation, we consider their distinctive characteristics as follows:

- Clear semantics: By explicitly incorporating four states—affirmation, contradictory negation, opposite negation, and intermediary negation—along with consistency constraints, the ambiguity of blank cells in classical FCA is eliminated.
- Rich expression: The intension of formal concepts simultaneously carries affirmation and three kinds of negation information, representing the four types of information: “what it is”, “what it is not”, “what it is opposite to”, and “what lies in between”. This enriches the granularity of knowledge representation, making knowledge units more complete.
- Logical coherence: Based on the set SCOI and the logic LCOI+PLCOI, SCOI and LCOI+PLCOI ensure semantic consistency as well as the soundness and completeness of reasoning.
- Enhanced expressiveness: It improves FCA’s capability to uncover complex relationships in knowledge and data, opening a whole new research dimension for FCA.

## 5. Attribute implication reasoning in FCACOI

In Formal Concept Analysis (FCA), one of the key focuses of reasoning is to discover implication relationships among attributes from the formal context (the relation between objects and attributes). These implications take the form of rules stating, "If an object possesses attribute set A, then it must also possess attribute set B." Such attribute implication reasoning describes the logical constraints among attributes and forms the foundation for understanding data features and constructing knowledge models. Therefore, attribute implication reasoning is the primary and most fundamental form of logical reasoning in FCA. It plays a role in revealing inherent logical dependencies among attributes and finds applications across many important fields including data science, knowledge engineering, databases, software engineering, and artificial intelligence. It serves as a fundamental tool for the logical organization, reduction, and automated reasoning of data and knowledge [13,19].

In FCA, the attribute implications are expressions of the form $A \to B$ with $A, B \subseteq M$, where M is the attribute set of a formal context. Such implications are valid if and only if every object having all attributes in A also has all attributes in B [5, 6]. Besides attribute implication reasoning, FCA also includes other reasoning forms such as fuzzy attribute implications [20], attribute exploration [21], concept reasoning [22], reduction reasoning [23] and composite reasoning methods combined with other logical frameworks [24]. These different forms of reasoning respectively take classical binary logic and non-classical logic as the logical basis of reasoning.

In this section, we mainly discuss the attribute implication reasoning in the formal concept analysis FCACOI. Among them, the logical foundation of attribute implication reasoning in FCACOI was discussed, as well as its application in a specific example.

## 5.1 Logical foundation of attribute implication reasoning in FCACOI

For the attribute implication reasoning in formal concept analysis FCACOI, based on the semantics of the logic LCOI+PLCOI, we introduce the notion of "ICOI-entailment" as the "semantic implication" for attribute implication reasoning in FCACOI. Through ICOI-entailment, a connection is established between attribute implication reasoning in FCACOI and inference in the logic LCOI+PLCOI. This demonstrates that formally proven inference rules in LCOI+PLCOI are valid in the attribute implication reasoning of FCACOI, and that LCOI+PLCOI provides a logical foundation for attribute implication reasoning in FCACOI.

**Definition 1** (ICOI-entailment). Let $S$ and $T$ ($S \neq T$) be attribute of objects (or set of attributes) in the formal context FCCOI with three types of negation. $S$ entails $T$ (i.e., every object possessing all attributes in $S$ also possesses all attributes in $T$), denoted as $S \models T$, if and only if every interpretation that satisfies $S$ also satisfies $T$ in the semantic interpretation of the logic LCOI+PLCOI. This kind of entailment is called "semantic implication" in attribute implication reasoning of FCACOI, abbreviated as ICOI-entailment.

In brief, the meaning of $S \models T$ in the attribute implication reasoning of FCACOI is that $T$ is a semantic consequence of $S$.

Concise speaking, in the attribute implication reasoning of FCACOI, $S \models T$ means that $T$ is a semantic consequence of $S$ (i.e., $S$ is the condition of semantic inference, and $T$ is the conclusion).

In classical FCA, an attribute of objects in a formal context is usually a statement or concept, which can be represented by an atomic formula in binary logic, and a set of attributes, corresponds to a set of formulas.

Similarly, since the formal context FCCOI is defined based on the set SCOI and the logic LCOI+PLCOI, an attribute in FCCOI corresponds to an atomic formula in LCOI+PLCOI, and a set of attributes corresponds to a set of formulas in LCOI+PLCOI. Therefore, $S$ and $T$ in Definition 1 correspond to a formula (or set of formulas) in LCOI+PLCOI. Based on the soundness theorem of LCOI+PLCOI (Theorem 1 in Section 3.2.2), for the axioms, deduction rules, and proven formal theorems in LCOI+PLCOI, $S \models T$ holds. Consequently, for the attribute implication reasoning in FCACOI, we obtain the following conclusions according to TCOI-entailment:

(1) If $T$ is an axiom in LCOI+PLCOI, then $\models T$.

(2) If $S$ and $T$ are respectively the premise and conclusion of a deduction rule in LCOI+PLCOI (i.e., $S \vdash T$), then $S \models T$.

(3) If $T$ is conclusion of a proven formal theorem in LCOI+PLCOI (i.e., $\vdash T$), then $\models T$.

(4) If $S$ and $T$ are respectively the premise and conclusion of a proven formal theorem in LCOI+PLCOI (i.e., $S \vdash T$), then $S \models T$.

Therefore, the attribute implication reasoning in FCACOI has the following reasoning properties:

**Property 1**. If $T$ is axiom in the logic LCOI+PLCOI, then

╞ A→(B→A); (A→(A→B))→(A→B); (A→B)→((B→C)→(A→C)); (A→¬B)→(B→¬A); (A→ ╕B)→(B→ ╕A); ¬A→(A→B); ((A→¬A)→B)→((A→B)→B); A→A∨B; B→A∨B; A∧B→A; A∧B→B; ╕A→ ¬A ∧ ¬~A; ~A→ ¬A ∧¬ ╕A

╞ ∀*x*A(*x*)→A(*a*); A(*a*)→∃*x*A(*x*); ∃*x*(A(*x*)→B)→(A(*a*)→B); ╕∀*x*A(*x*)→∃*x*╕A(*x*); ∃*x*╕A(*x*)→ ╕∀*x*A(*x*); ╕∃*x*A(*x*)→∀*x*╕A(*x*); ∀*x*╕A(*x*)→ ╕∃*x*A(*x*)

(Note. remove "╞", which is the axioms in LCOI+PLCOI (the axioms a1- a18 in Section 3.2.1))

**Property 2**. If $S$ and $T$ are respectively the premises and conclusions of the deduction rules in LCOI+PLCOI (i.e., $S \vdash T$), then

(1) $A_1, A_2, ..., A_n \models A_i$ ($i \in \{1, 2, ..., n\}$).

(2) $A \to B, A \models B$.

(3) If $A_1, A_2, ..., A_n \models A(a)$, where $a$ does not occur in $A_i$ ($1 \leq i \leq n$), then $A_1, A_2, ..., A_n \models \forall x A(x)$.

(Note. replace $\models$ with $\vdash$, that is the deduction rules of LCOI+PLCOI (D1, D2 and D3 in Section 3.2.1)

**Property 3**. If $T$ is conclusions of the proven formal theorems in LCOI+PLCOI (i.e., $\vdash T$), then

[1] ╞ A→A; ((A→B)→C)→(B→C); A→((A→B)→(C→B)); A→((A→B)→B)

[2] ╞ (((A→B)→B)→C)→(A→C); (A→(B→C))→(B→(A→C)); (B→C)→((A→B)→(A→C))

[3] ╞ ((A→B)→(A→(A→C)))→((A→B)→(A→C))

[4] ╞ (B→(A→C))→((A→B)→(A→C)); (A→(B→C))→((A→B)→(A→C))

[5] ╞ ¬A→(A→ ¬(B→B)); B→((A→ ¬B)→¬A); (A→ ¬B)→((A→B)→ ¬A)

[6] ╞ (A→B)→((A→ ¬B)→ ¬A); (A→B)→(¬B→ ¬A); (A→ ¬A)→ ¬A; A→ ¬ ¬A

[7] ╞ (¬A→ ¬B)→(B→A); ¬ ¬A→(¬ ¬A→A); ¬ ¬A→A; (¬A→B)→((¬A→ ¬B)→A)

[8] ╞ ╕A→(A→ ╕(B→B)); B→((A→ ╕B)→ ╕A); (A→ ╕B)→(A→(B→ ╕(B→B)))

[9] ╞ (A→B)→((A→ ╕B)→ ╕A); (A→B)→(╕B→ ╕A); (A→ ╕A)→ ╕A; A→ ╕╕A

[10] ╞ (╕A→ ╕B)→(B→A); ╕╕A→(╕╕A→A); ╕╕A→A; ¬(A∧¬A); ¬(╕A∧~A)

[11] ╞ ¬(A∧~A); ¬(A∧╕A)

(Note. replace ╞ with ├, that is the formal theorems in LCOI+PLCOI (theorem 1-theorem 4, Section 5.2.1 in [9]))

**Property 4**. If *S* and *T* are respectively the premises and conclusions of the proven formal theorems in LCOI+PLCOI (i.e., *S* ├ *T*), then

(1) A ╞ A; ╕╕A; ¬¬A; ¬╕A ∧ ¬~A; ¬╕~A

(2) ╕╕A ╞ A; ¬ ¬A ╞ A

(3) A, ╕B ╞ ╕(A→B); ╕(A→B) ╞ A, ╕B

(4) ~A ╞ ~╕A; ~╕A ╞ ~A

(5) ¬╕A ∧ ¬~A ╞ A

(6) A, ¬A ╞ B; A, ╕A ╞ B; A, ~A ╞ B

(Note. replace ╞ with ├, that is the formal theorems in LCOI+PLCOI (theorem 5 and theorem 6, Section 5.2.1 in [9]))

The above properties of attribute implication reasoning in FCACOI demonstrate that the formally proven inference rules in the logic LCOI+PLCOI are valid in the attribute implication reasoning of FCACOI. It can be stated that the attribute implication reasoning in FCACOI is equivalent to the inferences in the logic LCOI+PLCOI. This equivalence has the following significances:

- The logic LCOI+PLCOI establishes the logical foundation for the attribute implication reasoning in the formal concept analysis FCACOI with three kinds of negation, the attribute implication reasoning in FCACOI is logically self-consistent.

- For any formal deduce *S* ├ *T* in logic LCOI+PLCOI, according to the soundness theorem of LCOI+PLCOI (Theorem 1 in Section 3.2.2), *S* ╞ *T* holds for the attribute implication reasoning in FCACOI. That is, *S* ├ *T* ⇒ *S* ╞ *T*. Conversely, *S* ╞ *T* ⇒ *S* ├ *T* according to the completeness theorem [8, 9].

- FCACOI has stronger reasoning capabilities than FCA and its extensions.

- It deepens the integration of logical theory and formal concept analysis theory, expanding the boundaries of formal concept analysis in data analysis and knowledge processing.

## 5.2 Application of attribute implication reasoning in FCACOI

In traditional Chinese culture, the description of a person's life condition includes three dimensions: Health, Vitality and Spirit. Health (Jing) refers to the physical material foundation, meaning the body is free from illness and physiological functions are normal. Vitality (Qi) refers to the energy that drives bodily functions, manifested as agility and tirelessness. Spirit (Shen) refers to the external consciousness and radiance, shown by bright eyes and clear thinking. Describing a person's physical state with the unified concept of “Jing, Qi, Shen” represents the highest evaluation of a person's overall life energy fulfillment. Therefore, we can consider “Health”, “Vitality” and “Spirit” as the basic (original) attributes of a person.

In this section, we take “Health”, “Vitality” and “Spirit” as the original attributes of a person, and apply attribute implication reasoning in the formal concept analysis FCACOI to comprehensively assess a person's physical state.

Let the object set $U = \{x_1, x_2, x_3, x_4, x_5\}$ (representing five individuals), and the original attribute set $A = \{\text{Health } (A_1)\}$, Vitality ($A_2$), Spirit ($A_3$)} of objects in $U$.

(1) *Construction of the formal context*

Based on the set SCOI with three types of negation, the original attributes $A_1$, $A_2$, $A_3$ and their three types of negation representations and semantics are shown in the following table (Table 3).

**Table 3**. Original attributes and their three types of negation representations and semantics

| Negations / Attributes | Contradictory negation | Opposite negtion | Intermediary negation |
|---|---|---|---|
| Health ($A_1$) | Unhealthy ($A_1^{\neg}$) | Ill ($A_1^{ꓶ}$) | Sub-health ($A_1^{\sim}$)① |
| Vitality ($A_2$) | No Vitality ($A_2^{\neg}$) | Fatigue ($A_2^{ꓶ}$) | Ordinary ($A_2^{\sim}$)② |
| Spirit ($A_3$) | No Spirit ($A_3^{\neg}$) | Listless ($A_3^{ꓶ}$) | Ordinary ($A_3^{\sim}$)③ |

Among them, ① indicates 'neither unhealthy nor ill'; ② indicates 'neither without vitality nor fatigued'; ③ indicates 'neither spiritless nor listless'.

Obviously, the original attributes $A_1$, $A_2$, $A_3$ and their three kinds of negation are different fuzzy concepts (fuzzy sets). Since $A_1$, $A_2$ and $A_3$ are fuzzy sets, in fuzzy mathematics the membership degrees of objects to $A_1$, $A_2$ and $A_3$ can be obtained through expert opinions or statistical methods.

Assume that the membership degrees of objects $x_1$, $x_2$, $x_3$ and $x_4$ to the sets $A_1$, $A_2$ and $A_3$ (i.e., the degrees to which objects possess these attributes) are shown in the following table (Table 4):

**Table 4**. The degrees to which all objects possess the original attributes

| Attributes / Objects | $A_1$ | $A_2$ | $A_3$ |
|---|---|---|---|
| $x_1$ | 0.9 | 0.2 | 0.9 |
| $x_2$ | 0.7 | 0.7 | 0.6 |
| $x_3$ | 0.3 | 0.4 | 0.3 |
| $x_4$ | 0.1 | 0.8 | 0.2 |

According to the definition of the set SCOI (Definition 1 in Section 3.1), the membership degrees of each object to the contradictory negations $A_1^{\neg}$, $A_2^{\neg}$ and $A_3^{\neg}$, the opposite negations $A_1^{ꓶ}$, $A_2^{ꓶ}$ and $A_3^{ꓶ}$, and the intermediary negations $A_1^{\sim}$, $A_2^{\sim}$ and $A_3^{\sim}$ of attributes $A_1$, $A_2$ and $A_3$ can be determined.

For the sake of discussion, in the definition of the set SCOI, assume that $\lambda = 0.6$. Therefore, based on Table 4 and the definition of SCOI, the membership degree of object $x_1$ to the opposite negation $A_i^{ꓶ}$ of $A_i$ ($i$ =1, 2, 3, 4) can be calculated as $A_i^{ꓶ}(x_1) = 1–A_i(x_1) = 1– 0.9 = 0.1$; the membership degree of $x_1$ to the intermediary negation $A_i^{\sim}$ of $A_i$ is $A_i^{\sim}(x_1) = 0.45$; the membership degree of $x_1$ to the contradictory negation $A_i^{\neg}$ of $A_i$ is $A_i^{\neg}(x_1) = max(A_i^{ꓶ}(x_1), A_i^{\sim}(x_1)) = 0.45$. Similarly, the membership degrees $A_i^{ꓶ}(x_j)$, $A_i^{\sim}(x_j)$ and $A_i^{\neg}(x_j)$ for the other objects $x_j$ ($j$ = 2, 3, 4) listed in Table 4 can be computed accordingly. This results in Table 5.

Based on the above, a formal context $\mathbb{K}$ with three types of negation (Table 5) can be generated. $\mathbb{K}$ : ($U$, $A_{total}$, $I$), $U = \{x_1, x_2, x_3, x_4\}$, $A_{total} = A\cup A^{\neg}\cup A^{ꓶ}\cup A^{\sim}$, and $A = \{A_1, A_2, A_3\}$ is set of the original attributes. The values in the table represent the degree to which an object possesses an attribute. For example, $I(x_1, A_1) = 0.9$ indicates that object $x_1$ possesses attribute $A_1$ to the degree of 0.9.

**Table 5**. $\mathbb{K}$: formal context with three types of negation

| $U$ \ $A_{total}$ | $A\cup A^{\neg}\cup A^{\urcorner}\cup A^{\sim}$ | | | | | | | | | | | |
|---|---|---|---|---|---|---|---|---|---|---|---|---|
| | $A_1$ | $A_1^{\neg}$ | $A_1^{\urcorner}$ | $A_1^{\sim}$ | $A_2$ | $A_2^{\neg}$ | $A_2^{\urcorner}$ | $A_2^{\sim}$ | $A_3$ | $A_3^{\neg}$ | $A_3^{\urcorner}$ | $A_3^{\sim}$ |
| $x_1$ | 0.9 | 0.45 | 0.1 | 0.45 | 0.2 | 0.8 | 0.8 | 0.5 | 0.9 | 0.45 | 0.1 | 0.45 |
| $x_2$ | 0.7 | 0.55 | 0.3 | 0.55 | 0.7 | 0.55 | 0.3 | 0.55 | 0.6 | 0.6 | 0.4 | 0.6 |
| $x_3$ | 0.3 | 0.7 | 0.7 | 0.45 | 0.4 | 0.6 | 0.6 | 0.4 | 0.3 | 0.55 | 0.7 | 0.55 |
| $x_4$ | 0.1 | 0.9 | 0.9 | 0.55 | 0.8 | 0.5 | 0.2 | 0.5 | 0.2 | 0.8 | 0.8 | 0.5 |

(2) *Generate formal concepts and concept lattices with three types of negation*

All formal concepts derivable from the formal context $\mathbb{K}$ are shown in Table 6. Among them, the numerical values (decimals) following an attribute indicate the degree to which an object possesses that attribute.

**Table 6**. All formal concepts derivable from the formal context $\mathbb{K}$

| formal concepts | (Extension, Intension) |
|---|---|
| $C_1$ | $(\{x_1, x_2, x_3, x_4\}, \{A_{1\ 0.1}, A_{1\ 0.45}^{\neg}, A_{1\ 0.1}^{\urcorner}, A_{1\ 0.45}^{\sim}, A_{2\ 0.2}, A_{2\ 0.5}^{\neg}, A_{2\ 0.2}^{\urcorner}, A_{2\ 0.4}^{\sim}, A_{3\ 0.2}, A_{3\ 0.45}^{\neg}, A_{3\ 0.1}^{\urcorner}, A_{3\ 0.45}^{\sim}\})$ |
| $C_2$ | $(\{x_1, x_2, x_3\}, \{A_{1\ 0.3}, A_{1\ 0.45}^{\neg}, A_{1\ 0.1}^{\urcorner}, A_{1\ 0.45}^{\sim}, A_{2\ 0.2}, A_{2\ 0.55}^{\neg}, A_{2\ 0.3}^{\urcorner}, A_{2\ 0.4}^{\sim}, A_{3\ 0.3}, A_{3\ 0.45}^{\neg}, A_{3\ 0.1}^{\urcorner}, A_{3\ 0.45}^{\sim}\})$ |
| $C_3$ | $(\{x_1, x_2, x_4\}, \{A_{1\ 0.1}, A_{1\ 0.45}^{\neg}, A_{1\ 0.1}^{\urcorner}, A_{1\ 0.45}^{\sim}, A_{2\ 0.2}, A_{2\ 0.5}^{\neg}, A_{2\ 0.2}^{\urcorner}, A_{2\ 0.5}^{\sim}, A_{3\ 0.2}, A_{3\ 0.45}^{\neg}, A_{3\ 0.1}^{\urcorner}, A_{3\ 0.45}^{\sim}\})$ |
| $C_4$ | $(\{x_2, x_3, x_4\}, \{A_{1\ 0.1}, A_{1\ 0.55}^{\neg}, A_{1\ 0.3}^{\urcorner}, A_{1\ 0.45}^{\sim}, A_{2\ 0.4}, A_{2\ 0.5}^{\neg}, A_{2\ 0.2}^{\urcorner}, A_{2\ 0.4}^{\sim}, A_{3\ 0.2}, A_{3\ 0.55}^{\neg}, A_{3\ 0.4}^{\urcorner}, A_{3\ 0.5}^{\sim}\})$ |
| $C_5$ | $(\{x_1, x_2\}, \{A_{1\ 0.7}, A_{1\ 0.45}^{\neg}, A_{1\ 0.1}^{\urcorner}, A_{1\ 0.45}^{\sim}, A_{2\ 0.2}, A_{2\ 0.55}^{\neg}, A_{2\ 0.3}^{\urcorner}, A_{2\ 0.5}^{\sim}, A_{3\ 0.6}, A_{3\ 0.45}^{\neg}, A_{3\ 0.1}^{\urcorner}, A_{3\ 0.45}^{\sim}\})$ |
| $C_6$ | $(\{x_1, x_3\}, \{A_{1\ 0.3}, A_{1\ 0.45}^{\neg}, A_{1\ 0.1}^{\urcorner}, A_{1\ 0.45}^{\sim}, A_{2\ 0.2}, A_{2\ 0.6}^{\neg}, A_{2\ 0.6}^{\urcorner}, A_{2\ 0.4}^{\sim}, A_{3\ 0.3}, A_{3\ 0.45}^{\neg}, A_{3\ 0.1}^{\urcorner}, A_{3\ 0.45}^{\sim}\})$ |
| $C_7$ | $(\{x_2, x_3\}, \{A_{1\ 0.3}, A_{1\ 0.55}^{\neg}, A_{1\ 0.3}^{\urcorner}, A_{1\ 0.45}^{\sim}, A_{2\ 0.4}, A_{2\ 0.55}^{\neg}, A_{2\ 0.3}^{\urcorner}, A_{2\ 0.4}^{\sim}, A_{3\ 0.3}, A_{3\ 0.55}^{\neg}, A_{3\ 0.4}^{\urcorner}, A_{3\ 0.55}^{\sim}\})$ |
| $C_8$ | $(\{x_2, x_4\}, \{A_{1\ 0.1}, A_{1\ 0.55}^{\neg}, A_{1\ 0.3}^{\urcorner}, A_{1\ 0.55}^{\sim}, A_{2\ 0.7}, A_{2\ 0.5}^{\neg}, A_{2\ 0.2}^{\urcorner}, A_{2\ 0.5}^{\sim}, A_{3\ 0.2}, A_{3\ 0.6}^{\neg}, A_{3\ 0.4}^{\urcorner}, A_{3\ 0.5}^{\sim}\})$ |
| $C_9$ | $(\{x_3, x_4\}, \{A_{1\ 0.1}, A_{1\ 0.7}^{\neg}, A_{1\ 0.7}^{\urcorner}, A_{1\ 0.45}^{\sim}, A_{2\ 0.4}, A_{2\ 0.5}^{\neg}, A_{2\ 0.2}^{\urcorner}, A_{2\ 0.4}^{\sim}, A_{3\ 0.2}, A_{3\ 0.55}^{\neg}, A_{3\ 0.7}^{\urcorner}, A_{3\ 0.5}^{\sim}\})$ |
| $C_{10}$ | $(\{x_1\}, \{A_{1\ 0.9}, A_{1\ 0.45}^{\neg}, A_{1\ 0.1}^{\urcorner}, A_{1\ 0.45}^{\sim}, A_{2\ 0.2}, A_{2\ 0.8}^{\neg}, A_{2\ 0.8}^{\urcorner}, A_{2\ 0.5}^{\sim}, A_{3\ 0.9}, A_{3\ 0.45}^{\neg}, A_{3\ 0.1}^{\urcorner}, A_{3\ 0.45}^{\sim}\})$ |
| $C_{11}$ | $(\{x_2\}, \{A_{1\ 0.7}, A_{1\ 0.55}^{\neg}, A_{1\ 0.3}^{\urcorner}, A_{1\ 0.55}^{\sim}, A_{2\ 0.7}, A_{2\ 0.55}^{\neg}, A_{2\ 0.3}^{\urcorner}, A_{2\ 0.55}^{\sim}, A_{3\ 0.6}, A_{3\ 0.6}^{\neg}, A_{3\ 0.4}^{\urcorner}, A_{3\ 0.6}^{\sim}\})$ |
| $C_{12}$ | $(\{x_3\}, \{A_{1\ 0.3}, A_{1\ 0.7}^{\neg}, A_{1\ 0.7}^{\urcorner}, A_{1\ 0.45}^{\sim}, A_{2\ 0.4}, A_{2\ 0.6}^{\neg}, A_{2\ 0.6}^{\urcorner}, A_{2\ 0.4}^{\sim}, A_{3\ 0.3}, A_{3\ 0.55}^{\neg}, A_{3\ 0.7}^{\urcorner}, A_{3\ 0.55}^{\sim}\})$ |
| $C_{13}$ | $(\{x_4\}, \{A_{1\ 0.1}, A_{1\ 0.9}^{\neg}, A_{1\ 0.9}^{\urcorner}, A_{1\ 0.55}^{\sim}, A_{2\ 0.8}, A_{2\ 0.5}^{\neg}, A_{2\ 0.2}^{\urcorner}, A_{2\ 0.5}^{\sim}, A_{3\ 0.2}, A_{3\ 0.8}^{\neg}, A_{3\ 0.8}^{\urcorner}, A_{3\ 0.5}^{\sim}\})$ |
| $C_{14}$ | $(\varnothing,\{A_{1\ 1.0}, A_{2\ 1.0}, A_{3\ 1.0}\})$ |

All formal concepts $C_1$ to $C_{14}$ with three types of negation form a concept lattice $\mathbb{L}$ with three types of negation under the partial order $\leq$. Its Hasse diagram (Figure 9) is shown as follows:

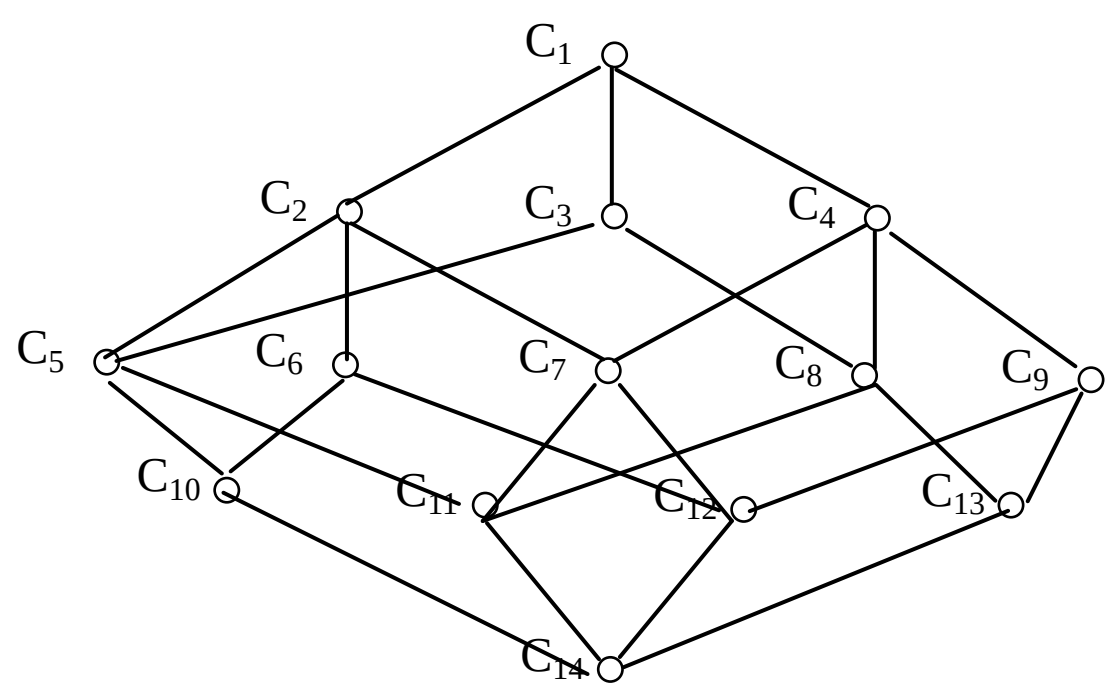

**Fig. 9**. Hasse diagram of concept lattice $\mathbb{L}$ with three types of negation

(3) *Extraction of attribute implication reasoning rules*

Based on the three types of negation (contradictory negation, opposite negation and intermediary negation) and their relations [8,9], we can extract the following attribute implication reasoning rules (values of the confidence levels are assumed and can be determined by domain experts' opinions or statistical methods).

1. Contradictory Negation-based reasoning rules (CNR):

CNR1: $A_1^{\neg} \rightarrow A_1^{\urcorner}$ (confidence 1.0). Meaning: If not healthy, then ill.

CNR2: $A_2^{\neg} \rightarrow A_2^{\urcorner}$ (confidence 0.7). Meaning: If no vitality, then fatigue.

CNR3: $A_3^{\neg} \rightarrow A_3^{\urcorner}$ (confidence 0.7). Meaning: If no spirit, then listless.

2. Opposite Negation-based reasoning rules (ONR):

ONR1: $A_1^{\urcorner} \rightarrow A_1^{\neg}$ (confidence 1.0). Meaning: If ill, then not healthy.

ONR2: $A_2^{\urcorner} \rightarrow A_2^{\neg}$ (confidence 0.8). Meaning: If fatigue, then no vitality.

ONR3: $A_3^{\urcorner} \rightarrow A_3^{\neg}$ (confidence 0.8). Meaning: If listless, then no spirit.

3. Intermediary Negation-based reasoning rules (INR):

INR1: $A_1^{\sim} \rightarrow A_1^{\neg} \cap A_1^{\urcorner \neg}$ (confidence 0.7). Meaning: If subhealthy, then neither healthy nor sick.

INR2: $A_2^{\sim} \rightarrow A_2^{\neg} \cap A_2^{\urcorner \neg}$ (confidence 0.7). Meaning: If vitality is average, then neither vitality nor fatigue.

INR3: $A_3^{\sim} \rightarrow A_3^{\neg} \cap A_3^{\urcorner \neg}$ (confidence 0.7). Meaning: If spirit is average, then neither no spirit nor listless.

*Remark*: Regarding rule 3, in the set SCOI, it has been proven that the intermediary negation $A^{\sim}$ of a set $A$ has the relationship with the contradictory negation $A^{\neg}$ and the opposite negation $A^{\urcorner}$ as follows: $A^{\sim} = A^{\neg} \cap A^{\urcorner \neg}$ (Property 3 in Section 5.2.1 of reference [9])

(4) *Attribute implication reasoning proxess*

Given observed data for a person $m$ on "Health", "Vitality" and "Spirit" as 0.4, 0.6, and 0.2 respectively, evaluate the physical condition of $m$.

The observed values 0.4, 0.6, and 0.2 represent the degrees (i.e., membership degrees) to which $m$ possesses attributes $A_1$, $A_2$ and $A_3$, i.e., $A_1(m) = 0.4$, $A_2(m) = 0.6$, $A_3(m) = 0.2$. Therefore, according to the definition of the set SCOI, the degrees (membership degrees) to which $m$ possesses the three kinds of negations of $A_1$, $A_2$ and $A_3$ can be calculated.

For the sake of brevity in the discussion, we place the degrees to which m possesses each attribute into the formal context K (Table 3), resulting in a comparison table (Table 7).

**Table 7**. Degree to which m possesses each attribute and the formal context $\mathbb{K}$

| $A_{total}$ / $U$ | $A \cup A^{\neg} \cup A^{\urcorner} \cup A^{\sim}$ | | | | | | | | | | | |
|---|---|---|---|---|---|---|---|---|---|---|---|---|
| | $A_1$ | $A_1^{\neg}$ | $A_1^{\urcorner}$ | $A_1^{\sim}$ | $A_2$ | $A_2^{\neg}$ | $A_2^{\urcorner}$ | $A_2^{\sim}$ | $A_3$ | $A_3^{\neg}$ | $A_3^{\urcorner}$ | $A_3^{\sim}$ |
| $x_1$ | 0.9 | 0.45 | 0.1 | 0.45 | 0.2 | 0.8 | 0.8 | 0.5 | 0.9 | 0.45 | 0.1 | 0.45 |
| $x_2$ | 0.7 | 0.55 | 0.3 | 0.55 | 0.7 | 0.55 | 0.3 | 0.55 | 0.6 | 0.6 | 0.4 | 0.6 |
| $x_3$ | 0.3 | 0.7 | 0.7 | 0.45 | 0.4 | 0.6 | 0.6 | 0.4 | 0.3 | 0.55 | 0.7 | 0.55 |
| $x_4$ | 0.1 | 0.9 | 0.9 | 0.55 | 0.8 | 0.5 | 0.2 | 0.5 | 0.2 | 0.8 | 0.8 | 0.5 |
| $m$ | **0.4** | 0.6 | 0.6 | 0.4 | **0.6** | 0.6 | 0.4 | 0.6 | **0.2** | 0.8 | 0.8 | 0.5 |

(5) *Conclusions*

As shown in Table 7, the maximum degree to which m possesses each attribute is 0.8, i.e., $I(m, A_3^{\neg}) = 0.8$ and $I(m, A_3^{\urcorner}) = 0.8$. Matching the known rules, the rules CNR3 and ONR3 will be triggered. Therefore, according to the inference rules CNR3 and ONR3, we derive the conclusion that m's physical condition is "No Spirit" and "Listless".

## 6. Research framework for attribute reduction of FCACOI

Attribute reduction in Formal Concept Analysis (FCA) aims primarily to identify and remove redundant attributes while preserving certain key system features, such as the concept structure and reasoning capabilities. Around this objective, a rich body of research has developed. Depending on the object and perspective of the reduction operation, two related yet distinct branches have emerged: attribute reduction of the formal context and attribute reduction of the concept lattice.

In FCA, attribute reduction of the formal context refers to the process of simplifying the attribute set in the formal context by removing redundant attributes. The goal of attribute reduction is to find a minimal subset of attributes, called a reduct, which can, in some sense, "represent" the entire set of attributes, preserving the ability to distinguish objects, the implication relations derived from attributes, and all other important conceptual information. In other words, the reduct retains the classification capability of the formal context as well as the logical structure among attributes. The concept lattice constructed over the reduct is isomorphic to the original concept lattice [6,19].

In this section, we discuss attribute reduction of the formal context FCCOI in FCACOI.

In FCACOI, a formal concept on a formal context is a quintuple ($X$, $B$, $C$, $O$, $D$), where the intension is no longer just a set of positive attributes, but consists of four component: the positive attribute set $B$, its contradictory negation $C$, opposite negation $O$ and intermediary negation $D$. Therefore, a reasonable attribute reduction for FCCOI should preserve the logical relationships of the formal concepts in all four intension components after reduction, maintain the implication relations derived from attributes, preserve the object-distinguishing ability, and ensure that the concept lattice constructed on the reduct is isomorphic to the concept lattice generated by the original formal context. To this end, we propose the following two research frameworks for attribute reduction of FCCOI based on different perspectives.

1. Attribute Reduction Preserving the Four Intension Components

A natural approach is to regard the formal context with three types of negation K = ($U$, $A_{total}$, $I$) as a "structural composition" of four sub-contexts $K_+$ = ($U$, $A$, $I$), $K_c$ = ($U$, $A^{\neg}$, $I$), $K_o$ = ($U$, $A^{⇁}$, $I$), $K_d$ = ($U$, $A^{\sim}$, $I$). Then, the attribute reduction for the formal context K can be transformed into performing attribute reduction on these four sub-contexts under the premise of preserving the consistency constraints of the four sub-contexts (i.e., mutual exclusion, Involution and completeness).

This type of reduction is theoretically ideal but also the strictest form of reduction. However, under the complex context with three types of negation, the computational complexity of finding such a reduction may be very high. We elaborate this attribute reduction framework for the concept lattice CLCOI from three perspectives: "structural composite", "reduction objectives" and "reduction methods".

(1) Structural composite.

In FCACOI, the four subcontexts $K_+$, $K_c$, $K_o$, $K_d$ of the formal context K share the same object set but have different attribute sets. The formal context K is not simply a straightforward union of the four individual contexts $K_+$, $K_c$, $K_o$ and $K_d$; rather, it should be regarded as a kind of "structural composition" of them. Under this composition, only those quintuples ($X$, $B$, $C$, $O$, $D$) that simultaneously satisfy the closure conditions, the four Galois connections and their properties, and are consistent with the set SCOI and logic LCOI+PLCOI are retained.

From the subcontext perspective, each subcontext $K_*$ ($* \in \{+, c, o, d\}$) generates a classical concept lattice $L_*$. In $L_*$, the set of objects $X$ ($X \subseteq U$) is the extension of the concept, and $B$, $C$, $O$, $D$ are the intensions and satisfying closure conditions, with the operators $*$ fulfilling the consistency constraints of K. Therefore, from a category theory perspective, the "structural composite" of the four classical concept lattices $L_*$ can be understood as their 'pullback' or fiber product (i.e., a limit construction in category theory of the common fibers of two morphisms) under the shared extension and consistency constraints. Such a mathematical construction of 'pullback' or 'fiber product' combines these lattices $L_*$ into a new structure that preserves the structure of the concept lattice CLCOI.

(2) Reduction objectives.

The goal of attribute reduction for the formal context K is to find the smallest subset of original attributes $A^{o} \subseteq A_{total}$ (i.e., an reduct) such that the concept lattice $CL_{COI}^{o}$ constructed based on $A^{o}$ through the aforementioned "structural composition' is isomorphic to the concept lattice $CL_{COI}$. This is equivalent to requiring that the four classical concept lattices $L_*$ satisfy the same consistency constraints after attribute reduction and produce a fiber product structure under these consistency constraints.

(3) Reduction method.

From subcontexts reduction to global reduction, the reduction steps are as follows:

Step 1: Subcontext reduction. For the four classical subcontexts $K_*$, similarly to attribute reduction methods in concept lattices (e.g., based on difference matrices, attribute characteristics, granular computing, etc.), find all their reducts.

Step 2: Imposing consistency constraints. Since the formal context $FC_{COI}$ with three types of negations must satisfy consistency constraints (mutual exclusivity, involution, completeness), the four Galois connection operators (Property 2 in Section 4.2) indicate that attributes have implication relations (contradictory negation implies opposite negation, contradictory implies intermediary) and mutual exclusivity. The implication relation requires that the reducts of the subcontexts satisfy "cross-context dependencies": if an attribute $a$ is retained in $K_o$, then $a$ must also be retained in $K_c$; if $a$ is retained in $K_d$, then $a$ must also be retained in $K_c$. Mutual exclusivity requires that the positive form and any one of the three negations cannot be retained simultaneously. Therefore, the reduct results must satisfy: among the four subcontexts, at most one retains the positive form $a$ (i.e., the original attribute). A reasonable constraint is that in the formal context $FC_{COI}$ with three types of negation (as in Table 1), if the $a^{ㄱ}$ column is retained, then the $a^{\neg}$ column must be retained; if the $a^{\sim}$ column is retained, then the $a^{\neg}$ column must be retained. Attributes $a$ and $a^{\neg}$ have no implication relation and can be retained independently. Completeness requires that each object has at most one of the attributes $a$, $a^{\neg}$, $a^{ㄱ}$ and $a^{\sim}$.

In summary, for the four subcontexts $K_*$ ($*\in\{+, c, o, d\}$), their consistency constraints can be transformed into dependency relationships among attribute sets; hence, their cross-context dependencies need to be considered.

Step 3: Global reduction. After obtaining the candidate reducts for each of the four subcontexts $K_*$, the above cross-context dependencies and consistency constraints are taken into account to determine a minimal attribute column set $S \subseteq A\cup A^{\neg}\cup A^{ㄱ}\cup A^{\sim}$ (Attribute columns: columns in the formal context that contain numbers, e.g., as shown in Table 1). How to find the minimal S (i.e., the minimal global reduct satisfying cross-context dependencies), this is similar to a multi-objective combinatorial optimization problem and can be solved by methods such as set cover and integer programming.

2. Attribute Reduction Preserving Attribute Implication Reasoning Capability

This is the most distinctive form of reduction in the framework we propose. Its goal is that for a formal context $FC_{COI}$, after attribute reduction, all formal inference rules (theorems) of the logic $L_{COI}$+$PL_{COI}$ that are valid in the original context remain valid. The reduced formal context retains the same attribute implication reasoning capability as the original formal context.

We will elaborate on the attribute reduction of the formal context $FC_{COI}$ from three aspects: "definition of attribute reduction preserving attribute implication reasoning", "object distinguishability" and "difference matrix-based reduction method".

(1) Definition of attribute reduction preserving attribute implication reasoning.

In Formal Concept Analysis $FCA_{COI}$, attribute implication reasoning takes the form $\Gamma\rightarrow\Delta$, where $\Gamma$ and $\Delta$ are sets of attributes of objects.. Therefore, we consider that an attribute reduction preserving the reasoning capability of the formal context $FC_{COI}$ means: the conclusions derived from attribute implication reasoning in the formal context FCCOI are completely consistent with those derived from attribute implication reasoning in the reduced formal context. In short, the attribute implication reasoning rules remain unchanged before and after the reduction of the formal context.

For a formal context K = ($U$, $A_{total}$, $I$) with three types of negation, the goal of attribute reduction is to find the minimal set of original attributes $A^{o} \subseteq A$ ($A$ is the set of original attributes in $A_{total}$). Therefore, for attribute reduction that preserves attribute implication reasoning, we can define it as follows:

Let $K_{A^{\circ}}$ denote the formal context with $A^{o}$ as the original attribute set (i.e., the reduced formal context). $K|_{A^{\circ}} \Rightarrow \varphi$ means that the conclusion $\varphi$ is derived from attribute implication reasoning in $K_{A^{\circ}}$, and $K \Rightarrow \varphi$ means that $\varphi$ is derived in K by attribute implication reasoning. If

$$K_{A^{\circ}} \Rightarrow \varphi, \text{ if and only if } K \Rightarrow \varphi$$

then the attribute reduction with $A^{o}$ as the original attribute set in the formal context is called an "attribute reduction preserving attribute implication reasoning", and $A^{o}$ is called the "attribute reduct set preserving attribute implication reasoning" in the formal context K.

(2) Object distinguishability.

In FCA, two objects are distinguishable on an attribute if and only if one object possesses the attribute while the other does not. For attribute reduction that preserves attribute implication reasoning in the formal context FCCOI, besides the above logical definition, preserving reasoning capability also requires that any pair of distinguishable objects in the original context remain distinguishable after reduction. Therefore, preserving reasoning capability must also satisfy: "the attribute reduct set $A^{o}$ can distinguish all distinguishable object pairs in the original formal context K". If two objects are indistinguishable on $A^{o}$, then they are also indistinguishable in any attribute implication reasoning involving attributes in $A^{o}$, which leads to failure of attribute implication inference.

From the formal context FCCOI with three types of negation (Table 1), two objects are distinguishable on an attribute as long as they differ in their degree of possessing that attribute; in other words, they can be distinguished by that attribute if their attribute values differ. Generally, for object distinguishability, we can define it as follows:

Let $x_1, x_2 \in U$, and mapping $\psi_{x \in U}: A^{\circ} \rightarrow \{+, \neg, ╕, \sim\}$. Objects $x_1$ and $x_2$ are distinguishable on the attribute reduct set $A^{\circ}$ if and only if there exists some attribute $m \in A^{\circ}$ such that $\psi_{x1}(m) \neq \psi_{x2}(m)$. Here, $\psi_{x1}(m) = +$ indicates object $x_1$ possesses attribute $m$, $\psi_{x1}(m) = \neg$ denotes that $x_1$ has the contradictory negation $m^{\neg}$ of $m$, $\psi_{x1}(m) =$ ╕ denotes $x_1$ has the opposite negation $m^{╕}$ of $m$, and $\psi_{x1}(m) = \sim$ denotes $x_1$ has the intermediary negation $m^{\sim}$ of $m$. Similarly, for $\psi_{x2}(m)$.

Based on the above, we conclude the following: For any two distinct objects $x_1, x_2 \in U$, if there exists an attribute $m \in A$ (where $A$ is the original attribute set in $A_{total}$) such that $\psi_{x1}(m) \neq \psi_{x2}(m)$, then there also exists an attribute $m' \in A^{\circ}$ such that $\psi_{x1}(m') \neq \psi_{x2}(m')$. In other words, the attribute reduct set $A^{\circ}$ must preserve the distinguishability of the original attribute set $A$ to the object set. This is precisely a direct generalization of the "preserving the extension of concept" condition in the attribute reduction of classical concept lattices.

(3) Reduction method based on the difference matrix.

In the formal context FCCOI with three types of negation, an original attribute $m \in A$ has three negations: $m^{\neg}$ (contradictory negation), $m^{╕}$ (opposite negation) and $m^{\sim}$ (intermediary negation). The degree to which an object $x \in U$ possesses one of these negated attributes is determined by the degree to which $x$ possesses the original attribute $m$. Therefore, the distinguishability of two objects with respect to the three negated attributes $m^{\neg}$, $m^{╕}$, $m^{\sim}$ can be reduced to their distinguishability with respect to the original attribute $m$. Thus, for attribute reduction of the formal context FCCOI, we can refer to the classical "difference matrix" method by constructing a difference matrix for object pairs $(x_i, x_j)$ to find a minimal attribute subset from the original attribute $A$. The specific reduction method is as follows:

Difference matrix: Let $T$ be a symmetric matrix of size $|U| \times |U|$, If the set of elements at position $(i, j)$ (for $i < j$) that is

$$D_{ij} = \{m \in A \mid \psi_{xi}(m) \neq \psi_{xj}(m)\},$$

$T$ is called the difference matrix. It contains all the original attributes that can distinguish between objects $x_i$ and

$x_j$.

Reduction condition: An attribute subset $A°$ ($A° \subseteq A$) is an attribute reduct set if and only if for every non-empty difference set $D_{ij}$, the intersection $A° \cap D_{ij} \neq \varnothing$. That is, $A°$ must have a non-empty intersection with every $D_{ij}$.

Minimal reduct: Find a minimal attribute reduct set $A°$ satisfying the above condition, this is a classical set covering problem and can be solved by Boolean logic or heuristic algorithms.

The above two attribute reduction research frameworks for the formal context FCCOI with three types of negation, their characteristics, and the similarities and differences with attribute reduction in classical formal context are summarized and compared below (Table 8).

**Table 8**. Comparison of two research frameworks for attribute reduction of the formal context FCCOI

| Dimension | Attribute Reduction in Classical Formal Context | Attribute Reduction Preserving Four Intension Components | Attribute Reduction Preserving Attribute Implication Reasoning Capability |
|---|---|---|---|
| *Core Objective* | Find a minimal subset of attributes such that the concept lattice constructed on this subset is isomorphic to the original lattice and the concept extens remain unchanged. | Under the premise of maintaining consistency constraints of four subcontexts, reduce the attributes of the four intension components. Find the minimal attribute subset such that the concept lattice based on it is isomorphic to the original concept lattice. | Preserving the formal inference rules (theorems) of the logic LCOI+PLCOI that are valid in the original formal context. The reduced formal context has the same attribute implication reasoning capability as the original formal context. |
| *Information Preservation* | Preserve the positive attribute set *B* and its correspondence with extension. | Preserve positive attribute *B* and its contradictory negation *C*, opposite negation *O*, and intermediary negation *D*. | The reduced concept lattice has the same logical consequences on the preserved attributes as before reduction. The attribute reduct set $A°$ can distinguish all distinguishable object pairs in the original formal context. |
| *Distinguishing Condition* | Two objects are distinguishable on attribute m if and only if one has m and the other does not. | Considering objects' attributes with four implicational components, two objects $x_i$ and $x_j$ are distinguishable on attribute m if and only if $\psi_{xi}(m) \neq \psi_{xj}(m)$, while satisfying cross-context constraints. | Two objects $x_i$ and $x_j$ are distinguishable on attribute m if and only if $\psi_{xi}(m) \neq \psi_{xj}(m)$ (without distinguishing specific negation types). |
| *Computation Method* | Difference matrix + set covering (or heuristic algorithms). | Construct difference matrices for four subcontexts respectively and solve for a global minimal attribute set satisfying cross-context dependencies. | Use classical difference matrix directly (only "different values" as the distinguishing condition) and solve set covering. |
| *Reduction Strength* | Medium (preserving positive information structure). | Strongest (preserving the structure of all four negation types) | Strong (preserving reasoning requires maintaining object distinguishability, but does not necessarily preserve concept lattice isomorphism). |
| *Logical Foundation* | Classical two-valued logic serves as the logical basis before and after attribute reduction of the formal context. | Logic LCOI+PLCOI serves as the logical basis before and after attribute reduction of the formal context, essentially a semantic model of LCOI+PLCOI. | Logic LCOI+PLCOI serves as the logical basis before and after attribute reduction of the formal context, preserving all inference rules in LCOI+PLCOI. |

**Common Poin:** All three construct difference matrices based on differences in object-attribute values, use set covering (or equivalent forms) to find minimal attribute subsets, and require the reduced formal context to distinguish all distinguishable object pairs from the original formal context (only the definition of "distinguishable" differs).
**Difference:** The attribute reduction of classical formal contexts deals only with binary (presence/absence) attributes. In the two types of attribute reduction for the formal context FCCOI with three kinds of negation, the attribute reduction that preserves the four concept extents needs to handle all four concept extents and maintain the complete relationships among them. The attribute reduction that preserves attribute implication reasoning must retain the attribute implication reasoning capability of the formal context and also maintain the distinguishability of objects.

For attribute reduction in FCA, attribute reduction of the formal context and attribute reduction of the concept lattice are the most fundamental and common approaches. The former compresses the data scale by selecting a subset of the original attributes, while the latter simplifies the knowledge representation by pruning the structure of the generated concept lattice. Together, they form the methodological foundation for attribute reduction in FCA. Therefore, for attribute reduction in FCACOI, in addition to the aforementioned "attribute reduction of the formal context FCCOI", there should also be attribute reduction methods such as "attribute reduction of the concept lattice CLCOI".

Regarding the attribute reduction of the formal context FCCOI and the attribute reduction of the concept lattice CLCOI, although their research objects and focuses differ, their goals are consistent: to simplify the attribute set and pursue a minimal attribute subset (attribute reduct) without losing the original distinguishing power and expressive capability. Since in formal concept analysis attribute reduction is usually first performed on the formal context and then the corresponding simplified concept lattice is constructed from the reduced formal context, it can be said that attribute reduction of the formal context FCCOI forms the basis for attribute reduction of the concept lattice CLCOI. Concerning the attribute reduction of the concept lattice CLCOI for formal concept analysis FCACOI with three kinds of negation, we can conduct research based on the above research frameworks.

## 7. Conclusions and future work

The classical Formal Concept Analysis (FCA) primarily focuses on the positive relationships between objects and attributes (i.e., "object has attribute") and lacks mechanisms to directly handle negation and its relations (such as "object does not have attribute"). To overcome this limitation, we introduce contradictory negation, opposite negation and intermediary negation into FCA. Based on the set SCOI and the logic LCOI+PLCOI, which incorporate these three types of negation, we define the formal context, Galois connection operators, formal concept and concept lattice with three kinds of negation. This leads to the proposal of a "Formal Concept Analysis with contradictory negation, opposite negation and intermediary negation (FCACOI)".

For the reasoning in FCACOI, this paper focuses on attribute implication reasoning. Based on the semantics of the logic LCOI+PLCOI, we introduce the notion of "ICOI-entailment" as the "semantic implication" for attribute implication reasoning in FCACOI. Through ICOI-entailment, a connection is established between attribute implication reasoning in FCACOI and inference in the logic LCOI+PLCOI, it indicate that formally proven inference rules (theorems) in LCOI+PLCOI are valid in the attribute implication reasoning of FCACOI, LCOI+PLCOI provides a logical foundation for attribute implication reasoning in FCACOI.

To illustrate the capability of attribute implication reasoning in FCACOI, we discuss its application in a concrete example. Moreover, we explore attribute reduction of the formal context in FCACOI, propose two research frameworks for attribute reduction from different perspectives, and compare their characteristics.

The Formal Concept Analysis FCACOI with contradictory negation, opposite negation and intermediary negation proposed in this paper is based on the set SCOI and the logic LCOI+PLCOI, which incorporate three types of negation. It takes the formal context FCCOI—with the original attribute set and its three types of negations as the attribute set—as the starting point, uses a quintuple of formal concepts as the basic knowledge unit, and employs the concept lattice CLCOI as the core data structure.

We believe that Formal Concept Analysis FCACOI with three types of negation is not only an extension of classical FCA that remedies its shortcomings in handling negation relations, but also a necessary response to the demands of complex negation semantics. Based on richer logic and semantics, FCACOI elevates FCA from a theory that “describes affirmations” to one that can “describe affirmations and its contradictions (either this or that), contraries (extreme negations) and intermediary (transitional states between contraries)”. FCACOI possesses adaptability and practical value in ensuring the integrity of knowledge representation, as well as in achieving finer-grained knowledge reduction and optimization, and it offers a new methodology for handling complex data that are imprecise, uncertain, polarized and gradual change.

Building on this paper, we will further study the theory and applications of FCACOI. For example, the integration of FCACOI with three-way decision theory, rough sets, non-monotonic reasoning, granular computing, knowledge graphs, and decision support systems; attribute reduction in formal contexts and concept lattices and their algorithms; and applications of FCACOI in information retrieval, data mining, and machine learning.